\PassOptionsToPackage{unicode}{hyperref}
\PassOptionsToPackage{hyphens}{url}
\PassOptionsToPackage{dvipsnames,svgnames,x11names}{xcolor}

\documentclass[12pt]{article}
\usepackage[utf8]{inputenc}
\usepackage{amsmath,amsthm,amssymb,amsfonts}

\usepackage{algorithm}
\usepackage{algorithmic}

\usepackage{enumitem}

\usepackage{indentfirst}
\usepackage{graphicx}
\graphicspath{{Figures/}}
\usepackage{subcaption}

\newtheorem{theorem}{Theorem}
\newtheorem{remark}{Remark}

\newtheorem{proposition}{Proposition}
\newtheorem{assumption}{Assumption}

\usepackage[section]{placeins}  
\DeclareFontFamily{U}{mathx}{}
\DeclareFontShape{U}{mathx}{m}{n}{<-> mathx10}{}
\DeclareSymbolFont{mathx}{U}{mathx}{m}{n}
\DeclareMathAccent{\widehat}{0}{mathx}{"70}
\DeclareMathAccent{\check}{0}{mathx}{"71}

\usepackage{iftex}
\ifPDFTeX
  \usepackage[T1]{fontenc}
  \usepackage[utf8]{inputenc}
  \usepackage{textcomp} % provide euro and other symbols
\else % if luatex or xetex
  \usepackage{unicode-math}
  \defaultfontfeatures{Scale=MatchLowercase}
  \defaultfontfeatures[\rmfamily]{Ligatures=TeX,Scale=1}
\fi
\usepackage{lmodern}
\ifPDFTeX\else  
\fi
\IfFileExists{upquote.sty}{\usepackage{upquote}}{}
\IfFileExists{microtype.sty}{% use microtype if available
  \usepackage[]{microtype}
  \UseMicrotypeSet[protrusion]{basicmath} % disable protrusion for tt fonts
}{}
\makeatletter
\@ifundefined{KOMAClassName}{% if non-KOMA class
  \IfFileExists{parskip.sty}{%
    \usepackage{parskip}
  }{% else
    \setlength{\parindent}{0pt}
    \setlength{\parskip}{6pt plus 2pt minus 1pt}}
}{% if KOMA class
  \KOMAoptions{parskip=half}}
\makeatother
\usepackage{xcolor}
\makeatletter
\ifx\paragraph\undefined\else
  \let\oldparagraph\paragraph
  \renewcommand{\paragraph}{
    \@ifstar
      \xxxParagraphStar
      \xxxParagraphNoStar
  }
  \newcommand{\xxxParagraphStar}[1]{\oldparagraph*{#1}\mbox{}}
  \newcommand{\xxxParagraphNoStar}[1]{\oldparagraph{#1}\mbox{}}
\fi
\ifx\subparagraph\undefined\else
  \let\oldsubparagraph\subparagraph
  \renewcommand{\subparagraph}{
    \@ifstar
      \xxxSubParagraphStar
      \xxxSubParagraphNoStar
  }
  \newcommand{\xxxSubParagraphStar}[1]{\oldsubparagraph*{#1}\mbox{}}
  \newcommand{\xxxSubParagraphNoStar}[1]{\oldsubparagraph{#1}\mbox{}}
\fi
\makeatother

\usepackage{longtable,booktabs,array}
\usepackage{calc} % for calculating minipage widths
\usepackage{etoolbox}
\makeatletter
\patchcmd\longtable{\par}{\if@noskipsec\mbox{}\fi\par}{}{}
\makeatother
\IfFileExists{footnotehyper.sty}{\usepackage{footnotehyper}}{\usepackage{footnote}}
\makesavenoteenv{longtable}
\usepackage{graphicx}
\makeatletter
\def\maxwidth{\ifdim\Gin@nat@width>\linewidth\linewidth\else\Gin@nat@width\fi}
\def\maxheight{\ifdim\Gin@nat@height>\textheight\textheight\else\Gin@nat@height\fi}
\makeatother
\setkeys{Gin}{width=\maxwidth,height=\maxheight,keepaspectratio}
\makeatletter
\def\fps@figure{htbp}
\makeatother

\makeatletter
\@ifpackageloaded{caption}{}{\usepackage{caption}}
\AtBeginDocument{%
\ifdefined\contentsname
  \renewcommand*\contentsname{Table of contents}
\else
  \newcommand\contentsname{Table of contents}
\fi
\ifdefined\listfigurename
  \renewcommand*\listfigurename{List of Figures}
\else
  \newcommand\listfigurename{List of Figures}
\fi
\ifdefined\listtablename
  \renewcommand*\listtablename{List of Tables}
\else
  \newcommand\listtablename{List of Tables}
\fi
\ifdefined\figurename
  \renewcommand*\figurename{Figure}
\else
  \newcommand\figurename{Figure}
\fi
\ifdefined\tablename
  \renewcommand*\tablename{Table}
\else
  \newcommand\tablename{Table}
\fi
}
\@ifpackageloaded{float}{}{\usepackage{float}}
\floatstyle{ruled}
\@ifundefined{c@chapter}{\newfloat{codelisting}{h}{lop}}{\newfloat{codelisting}{h}{lop}[chapter]}
\floatname{codelisting}{Listing}

\makeatother
\makeatletter
\@ifpackageloaded{caption}{}{\usepackage{caption}}
\@ifpackageloaded{subcaption}{}{\usepackage{subcaption}}
\makeatother

\ifLuaTeX
  \usepackage{selnolig}  % disable illegal ligatures
\fi
\usepackage[]{natbib}
\usepackage{bookmark}

\IfFileExists{xurl.sty}{\usepackage{xurl}}{} % add URL line breaks if available
\hypersetup{
  pdftitle={Title},
  pdfauthor={Author 1; Author 2},
  pdfkeywords={3 to 6 keywords, that do not appear in the title},
  colorlinks=true,
  linkcolor={blue},
  filecolor={Maroon},
  citecolor={Blue},
  urlcolor={Blue},
  pdfcreator={LaTeX via pandoc}}

\newcommand{\anon}{1}

\begin{document}
\begin{sloppypar}
\def\spacingset#1{\renewcommand{\baselinestretch}%
{#1}\small\normalsize} \spacingset{1}
%%%%%%%%%%%%%%%%%%%%%%%%%%%%%%%%%%%%%%%%%%%%%%%%%%%%%%%%%%%%%%%%%%%%%%%%%%%%%%

\if1\anon
{ 
  \title{\bf %Beyond Shared and Private: 
  Hierarchical Contrastive Learning for Multimodal Data}
  \author{Huichao Li \\
  %\thanks{
    %The authors gratefully acknowledge \textit{please remember to list all relevant funding sources in the version that gives all author information}}\hspace{.2cm}\\
    {\small Department of Mathematical Sciences, University of Chinese Academy of Sciences}\\
    Junhan Yu \\
    {\small Department of Statistics and Data Science, National University of Singapore }\\
    and \\
    Doudou Zhou\footnote{Corresponding author: ddzhou@nus.edu.sg} \\
    {\small Department of Statistics and Data Science, National University of Singapore} }
    \date{}
  \maketitle
} \fi

\if0\anon
{
  \bigskip
  \bigskip
  \bigskip
  \begin{center}
    {\LARGE\bf Hierarchical Contrastive Learning for Multimodal Data %Beyond Shared and Private: Hierarchical Contrastive Learning for Multimodal Data with Partial Sharing
    }
\end{center}
  \medskipglo
} \fi

\bigskip
\begin{abstract}
Multimodal representation learning is commonly built on a shared-private decomposition, treating latent information as either common to all modalities or specific to one. This binary view is often inadequate: many factors are shared by only subsets of modalities, and ignoring such partial sharing can over-align unrelated signals and obscure complementary information. We propose Hierarchical Contrastive Learning (HCL), a framework that learns globally shared, partially shared, and modality-specific representations within a unified model. HCL combines a hierarchical latent-variable formulation with structural sparsity and a structure-aware contrastive objective that aligns only modalities that genuinely share a latent factor. Under uncorrelated latent variables, we prove identifiability of the hierarchical decomposition, establish recovery guarantees for the loading matrices, and derive parameter estimation and excess-risk bounds for downstream prediction. Simulations show accurate recovery of hierarchical structure and effective selection of task-relevant components. On multimodal electronic health records, HCL yields more informative representations and consistently improves predictive performance.
\end{abstract}

\noindent%

\vfill

\newpage
\spacingset{1.8}

\section{Introduction}\label{sec:introduction}
Multimodal learning has become a central paradigm in modern machine learning because real-world entities are often observed through multiple heterogeneous views, such as text, images, and audio. Across these applications, a core objective is to learn low-dimensional representations that capture cross-modal relationships while preserving information relevant for retrieval, prediction, reasoning, and decision-making \citep{guo2019deep}. The success of such representations depends not only on aligning heterogeneous observations, but also on determining \emph{which} information is shared across all modalities, \emph{which} is shared only by subsets of modalities, and \emph{which} remains modality-specific \citep{baltruvsaitis2018multimodal}.

A common strategy in multimodal representation learning is to project unimodal features into a shared semantic space and then perform fusion within that space \citep{poria2016fusing,xia2023achieving,ng2025spaner,zhu2025unified}. Although effective for multimodal fusion, this strategy does not distinguish modality-specific information. To address this limitation, a growing body of work seeks to disentangle shared and modality-specific information, thereby improving robustness and discriminative performance \citep{hazarika2020misa,yao2024drfuse,wang2025dlf,zhang2026partially}. However, this shared-versus-private dichotomy remains too coarse once more than two modalities are present, because many latent factors may be shared by only subsets of modalities. For example, in vision-language-audio settings, some semantic information may be jointly encoded by images and text, whereas audio may capture only a distinct or weaker aspect of the same underlying signal \citep{yue2025coavt}. Ignoring such partially shared structure can lead to systematic representational errors, as signals may be inappropriately forced into a fully shared space or dismissed as modality-specific noise. This can produce over-alignment, under-utilization of complementary information, increased fragility to missing modalities, and reduced interpretability in downstream predictions. Partially shared structure should therefore be viewed as not a minor refinement of multiview decomposition, but a fundamental missing abstraction in multimodal representation learning.

Recent advances in contrastive learning have substantially accelerated progress in multimodal representation learning, but existing methods either rely on a coarse shared-versus-private decomposition or treat partial sharing in largely heuristic ways. Methods such as FACTORCL \citep{liang2023factorized}, QUEST \citep{song2024quest}, CoMM \citep{dufumier2024align}, and COrAL \citep{cissee2026orthogonalized} enrich the shared-private paradigm through factorization, orthogonality, or synergy-based designs. \cite{meng2026tri} moved beyond this dichotomy by introducing a tri-subspace disentanglement framework with partially shared components, but it remains an architecture-driven empirical approach and does not clearly quantify or interpret the contribution of each latent component in downstream prediction. 

In parallel, recent theory has clarified several aspects of multimodal contrastive representation learning. \cite{yao2023multi} established identifiability for arbitrary subsets of views in observational multiview causal representation learning with partial observability. \cite{cai2025value} studied cross-modal misalignment through a latent-variable model and showed that multimodal contrastive learning recovers semantic variables that are invariant to selection and perturbation biases. \cite{lin2025statistical} showed that SimCLR-type objectives \citep{chen2020simple} can yield approximately sufficient representations that adapt effectively to downstream regression and classification tasks. \cite{alvandi2025revisiting} derived generalization bounds for downstream supervised loss under distribution mismatch, including domain shift and domain generalization. However, these approaches generally do not provide a unified probabilistic framework for hierarchical sharing across arbitrary subsets of modalities. Existing theory focuses on fully shared factors, invariant coupled variables, or general transfer performance, and therefore does not offer a complete statistical treatment of identifiability, representation recovery, and selection of task-relevant latent components in downstream prediction. As a result, partially shared structure is often acknowledged empirically, but it is still not handled in a way that is both naturally integrated with representation learning and theoretically well grounded.

A related statistical literature studies partial sharing primarily from the perspective of multiview decomposition and structured subspace estimation rather than representation learning. \cite{gaynanova2019structural} introduced a deterministic matrix decomposition framework that explicitly incorporates globally shared, partially shared, and modality-specific structures. \cite{prothero2024data} further developed DIVAS, a subspace-based method that identifies partially shared components through sequential search and angular perturbation analysis. \cite{mao2026supervised} incorporated downstream supervision to identify predictive latent components and extended the framework to incomplete multimodal data. On the theoretical side, much of the existing analysis remains centered on recovery of the jointly shared subspace under the AJIVE framework \citep{feng2018angle}, which does not accommodate partially shared structure. \cite{yang2025estimating} showed that AJIVE achieves optimal rates in high signal-to-noise regimes but exhibits non-diminishing error in low signal-to-noise ratio settings. \cite{li2025heterojive} extended this framework by introducing weighted spectral aggregation for heterogeneous views, thereby reducing the bias term and establishing improved recovery rates under general geometric conditions. \cite{ma2026optimal} studied optimal estimation of shared singular subspaces across multiple noisy matrices, characterized the limitations of stacked singular value decomposition (SVD) under partial sharing, and proposed minimax-optimal estimators for both shared and unshared singular vectors. Although these works provide valuable structural insight, they are formulated mainly for deterministic matrices or structured subspace models, often under relatively strong geometric conditions. Moreover, the theory is directed mainly at jointly shared structure and does not offer a unified treatment of partially shared and modality-specific components. What remains missing is a framework that simultaneously models hierarchical sharing, supports representation learning, and provides statistical guarantees for latent-structure recovery and downstream prediction.

\begin{figure}[t]
    \centering
    \includegraphics[width=0.98\linewidth]{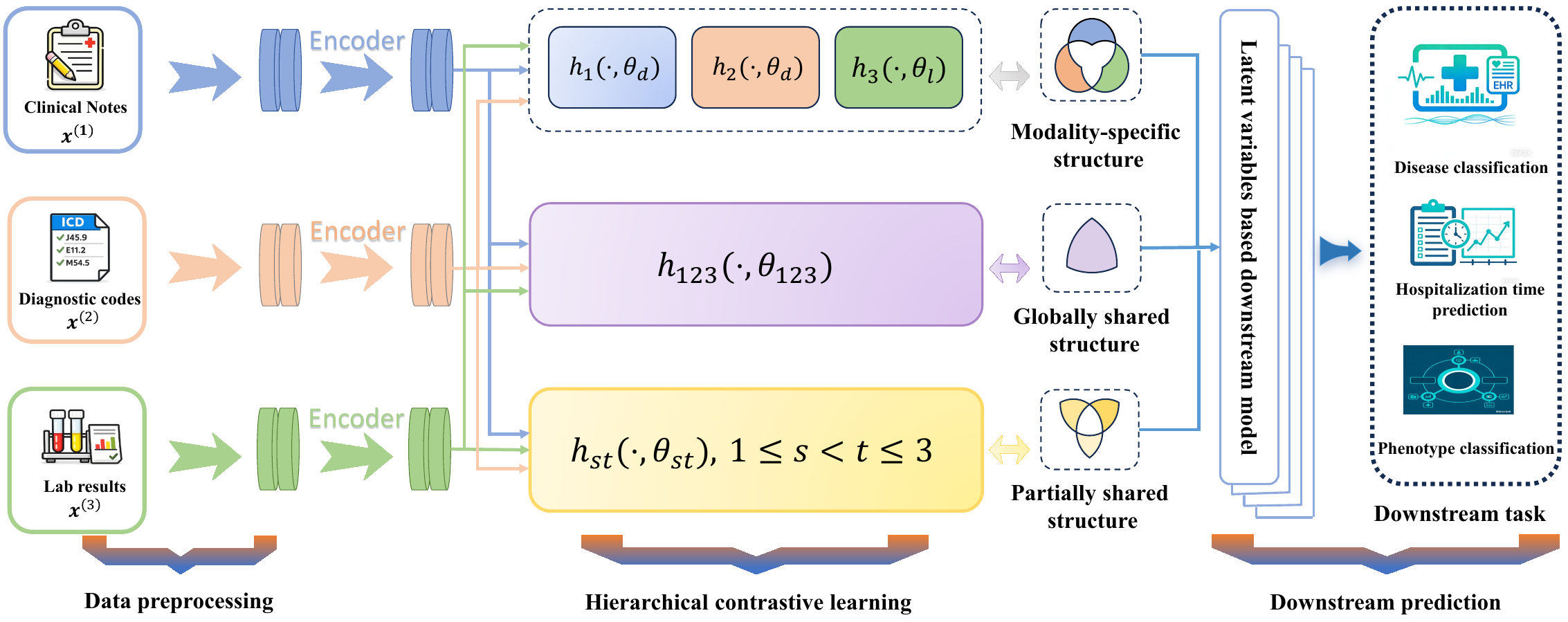}
    \caption{Overall framework of Hierarchical Contrastive Learning. In Hierarchical contrastive learning module, each encoder function $h$ is designed to recover a specific latent structure. The illustration uses electronic health record data in Section~\ref{sec:real_data} as a concrete example, but the framework extends directly to general multimodal data given appropriate encoders.}
    \label{fig:framework}
\end{figure}

To address these limitations, we propose \textbf{H}ierarchical \textbf{C}ontrastive \textbf{L}earning (HCL), a framework that treats multimodal representation learning as the problem of recovering hierarchically organized latent structure. Our contributions are threefold. First, we formulate a hierarchical latent-variable model for multimodal data that explicitly captures globally shared, partially shared, and modality-specific components, and we establish identifiability of the decomposition under mild assumptions. Second, building on this hierarchical decomposition, we propose a hierarchical contrastive objective for learning the corresponding representations and derive recovery guarantees for the estimated loading matrices. Third, we connect the learned representations to downstream prediction through debiased and group-regularized linear regression, deriving parameter estimation and excess-risk bounds. Taken together, these results provide a unified statistical framework for multimodal representation learning with hierarchical sharing. The architecture of HCL is illustrated in Figure~\ref{fig:framework}.

The remainder of the paper is organized as follows. Section~\ref{sec:method} introduces the proposed method and Section~\ref{sec:statistical property} presents its theoretical properties. Sections~\ref{sec:simulation} and \ref{sec:real_data} provide simulation and
real-world analysis. Section~\ref{sec:discussion} concludes with a discussion.

\section{Method}\label{sec:method}
\noindent \textbf{Notation.\,} 
For any positive integer $n$, let $[n]=\{1,2,\cdots,n\}$. For any matrix $\mathbf{A}$, let $\|\mathbf{A}\|$, $\|\mathbf{A}\|_{\rm F}$, $\|\mathbf{A}\|_{2,\infty}$ be the spectral norm, Frobenius norm and the largest $\ell_2$ norm of the rows of $\mathbf{A}$, respectively. Let $\sigma_j(\mathbf{A})$ be the $j$-th largest singular value of $\mathbf{A}$. Let $\mathcal{O}_{d,r}$ $(d\geq r)$ denote the set of $d\times r$ orthonormal matrices. For sequences $\{a_n\},\{b_n\}>0$, write $a_n\lesssim b_n$ or $a_n=O(b_n)$ if $a_n\leq C b_n$ for some $C>0$ and all $n$, and write $a_n\ll b_n$ if $a_n\leq b_n/C$ for a sufficiently large $C>0$, and $a_n \asymp b_n$ if $a_n\lesssim b_n$ and $b_n\lesssim a_n$. Denote $\langle \boldsymbol{a},\boldsymbol{b} \rangle=\boldsymbol{a}^\top \boldsymbol{b}$ for any vectors $\boldsymbol{a},\boldsymbol{b}\in\mathbb{R}^d$.

\subsection{Hierarchical Structure of Multimodal Data}
Suppose that we observe $n$ independent samples, each consisting of $M$ modalities. For the $i$-th sample, let $\boldsymbol{x}_i^{(m)} \in \mathcal{X}_m$ denote the random variable associated with modality $m$, where $\mathcal{X}_m$ is the feature space of modality $m$, and the complete observation is given by $\boldsymbol{x}_i = \{\boldsymbol{x}_i^{(1)},\cdots,\boldsymbol{x}_i^{(M)}\}$. To address the inherent heterogeneity of multimodal data, recent research in representation learning has focused on decomposing multimodal information into globally shared and modality-specific components. When more than two modalities are present, however, some information may be shared by only a subset of modalities, giving rise to partially shared structure. To capture these complex cross-modal dependencies, we introduce a hierarchical decomposition of the latent space into globally shared, partially shared, and modality-specific components.

\begin{figure}[t]
        \centering
        \subfloat[General hierarchical structure]{\raisebox{0.02\linewidth}{\includegraphics[width=0.38\linewidth]{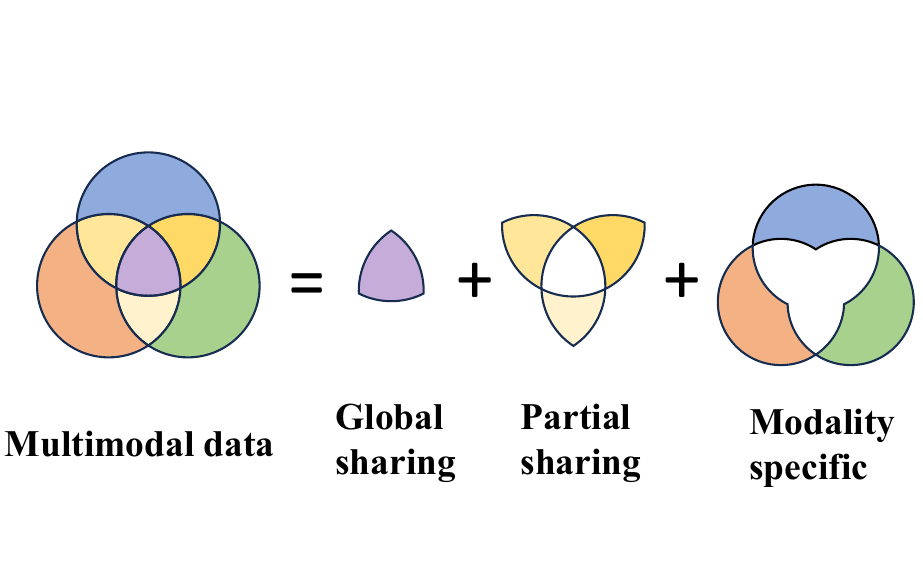}}\label{fig:general_structure}}
        \hspace{0.02\linewidth}
        \centering
        \subfloat[][Hierarchical decomposition model]{\includegraphics[width=0.58\linewidth]{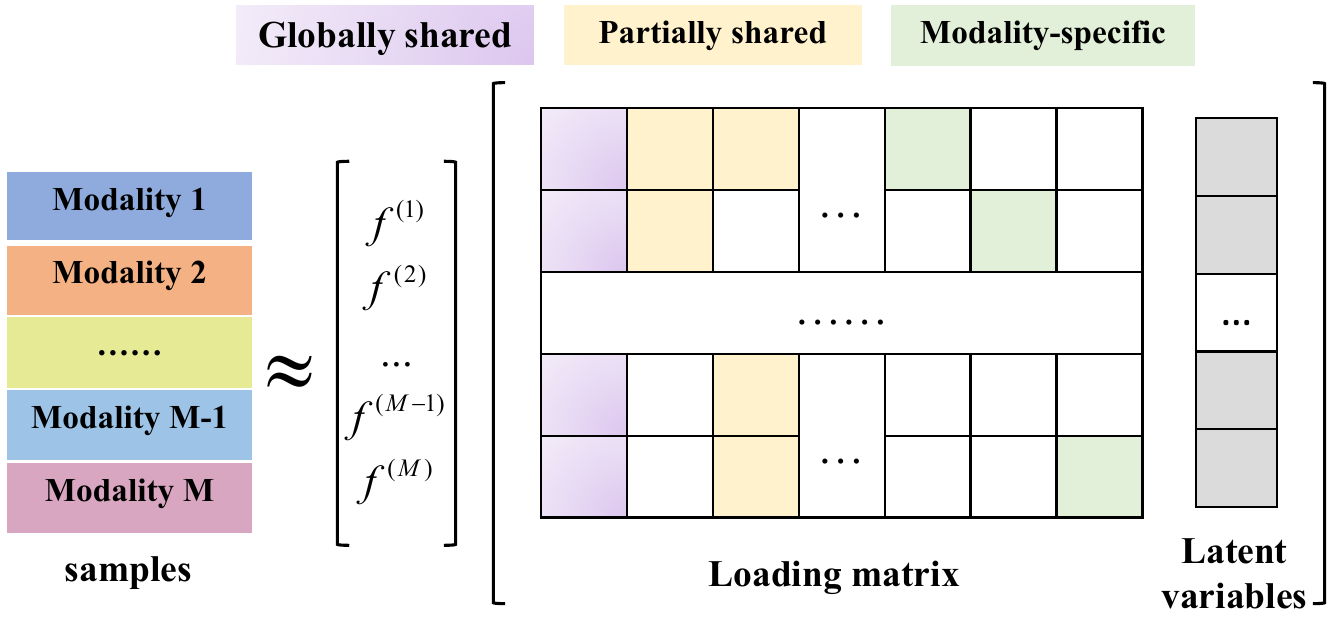}\label{fig:hierarchical_model}}
        \caption{Hierarchical decomposition for multimodal data.}
\end{figure}

We illustrate the proposed hierarchical decomposition with three modalities for clarity, though the framework readily generalizes to settings involving more than three modalities. Let the collection of latent structures be denoted by $S=S_1\cup S_2\cup S_3 $, where 
\begin{align*}
\text{Level 1 (Globally shared structure)}: S_1=&\left\{\{1,2,3\}\right\}, \\
\text{Level 2 (Partially shared structure)}: S_2=&\left\{\{1,2\},\{1,3\},\{2,3\}\right\}, \\
\text{Level 3 (Modality-specific structure)}: S_3=&\left\{\{1\},\{2\},\{3\}\right\}.
\end{align*}
Each element $s\in S$ corresponds to a latent structure shared by the modalities indexed by its entries. For example, $s=\{1,2\}$ represents the partially shared structure common to modality 1 and 2 only. Thus, Level 1 captures the structure shared across all three modalities, Level 2 captures pairwise shared structures, and Level 3 captures modality-specific structures. Figure~\ref{fig:general_structure} illustrates the hierarchical structure with three modalities.

It is worth noting that a full hierarchical decomposition is not always necessary in practice. For example, when partially shared structure is absent, the framework reduces to a simpler decomposition involving only globally shared and modality-specific components, as in existing approaches such as matrix factorization \citep{lock2013joint}, canonical correlation analysis \citep{shu2020d}, and factor regression \citep{li2022integrative}. Closely related to our formulation, \cite{gaynanova2019structural,yi2023hierarchical,prothero2024data} also considered the partially shared structure, although their focus was on deterministic matrix decomposition rather than representation learning.

\subsection{Hierarchical Decomposition Model}
To model the generative process underlying the proposed hierarchical structure, we assume the observed data for sample $i$ are generated as follows:
\begin{equation}\label{eq:model_hierarchical}
\begin{split}
\boldsymbol{x}_i^{(1)}&=f^{(1)}\big(\mathbf{W}_{123}^{(1)}\boldsymbol{z}_{123,i} + \mathbf{W}_{12}^{(1)}\boldsymbol{z}_{12,i} + \mathbf{W}_{13}^{(1)}\boldsymbol{z}_{13,i} + \mathbf{W}_{1}^{(1)}\boldsymbol{z}_{1,i} + \boldsymbol{\epsilon}_i^{(1)}\big), \\
\boldsymbol{x}_i^{(2)}&=f^{(2)}\big(\mathbf{W}_{123}^{(2)}\boldsymbol{z}_{123,i} + \mathbf{W}_{12}^{(2)}\boldsymbol{z}_{12,i} + \mathbf{W}_{23}^{(2)}\boldsymbol{z}_{23,i} + \mathbf{W}_{2}^{(2)}\boldsymbol{z}_{2,i} + \boldsymbol{\epsilon}_i^{(2)}\big), \\
\boldsymbol{x}_i^{(3)}&=f^{(3)}\big(\mathbf{W}_{123}^{(3)}\boldsymbol{z}_{123,i} + \mathbf{W}_{13}^{(3)}\boldsymbol{z}_{13,i} + \mathbf{W}_{23}^{(3)}\boldsymbol{z}_{23,i} + \mathbf{W}_{3}^{(3)}\boldsymbol{z}_{3,i} + \boldsymbol{\epsilon}_i^{(3)}\big), \\
\end{split}
\end{equation}
where $f^{(m)}:\mathbb{R}^{d_m}\rightarrow\mathcal{X}_m$ denotes the link function mapping the latent space to the observed space for modality $m$, $\boldsymbol{z}_{s,i} \in\mathbb{R}^{r_{s}}$ is the latent variable associated with sample $i$ and structure $s$, and $\mathbf{W}_{s}^{(m)}\in\mathcal{R}^{d_m\times r_{s}}$ is the true loading matrix for structure $s$ in modality $m$. Here, whenever no ambiguity arises, we simplify the notation by writing $\boldsymbol{z}_{\{1,2,3\},i}$ as $\boldsymbol{z}_{ 123,i}$ and $\mathbf{W}_{\{1,2,3\}}^{(1)}$ as $\mathbf{W}_{123}^{(1)}$, and analogously for the remaining structures.
For identifiability, we assume that the latent variables $\boldsymbol{z}_{s,i}$ have mean zero and identity covariance, and satisfy $\mathrm{Cov}(\boldsymbol{z}_{s,i},\boldsymbol{z}_{t,i})=0$ for $s\neq t$. We further assume that the error vector $\boldsymbol{\epsilon}_i=(\boldsymbol{\epsilon}_i^{(1)\top},\boldsymbol{\epsilon}_i^{(2)\top},\boldsymbol{\epsilon}_i^{(3)\top})^\top$ is independent of $\{\boldsymbol{z}_{s,i},s\in S\}$ with covariance $\sigma_{\epsilon}^2\mathbf{I}_d$, where $d=\sum_{m=1}^3d_m$. If structure $s$ is absent in modality $m$, the corresponding loading matrix $\mathbf{W}_{s}^{(m)}$ is set to zero. In addition, for each modality $m$, we assume that $\mathbf{W}_{\rm act}^{(m)}=[\mathbf{W}_{s}^{(m)}]_{m\in s,s\in S_*}$ has full column rank, where $S_*$ denotes the set of structures with nonzero loading matrices.  Figure~\ref{fig:hierarchical_model} illustrates the hierarchical decomposition model for an arbitrary number of modalities.

\begin{remark}
    In practice, features from different modalities often lie in distinct and incomparable spaces. For the data generated by model~\eqref{eq:model_hierarchical}, if the link functions $f^{(m)}$ are invertible, one may apply $(f^{(m)})^{-1}(\boldsymbol{x}_i^{(m)})$ to map observations from different modalities into a common semantic subspace, a strategy widely used in multimodal data integration \citep{poria2016fusing,xia2023achieving,zhu2025unified}. The proposed framework is also highly general. In particular, when the linear latent-factor component is interpreted as the final layer of a neural network and $f^{(m)}$ represents the preceding nonlinear layers, the model encompasses a broad class of neural network architectures. Moreover, we only need the latent variables of different structure are uncorrelated rather independent, allowing for potentially complex nonlinear dependence among them.
\end{remark}

Let the complete latent representation for sample $i$ be $$\boldsymbol{z}_i=\left(\boldsymbol{z}_{123,i}^\top,\boldsymbol{z}_{12,i}^\top,\boldsymbol{z}_{13,i}^\top,\boldsymbol{z}_{23,i}^\top,\boldsymbol{z}_{1,i}^\top,\boldsymbol{z}_{2,i}^\top,\boldsymbol{z}_{3,i}^\top\right)^\top\in\mathbb{R}^{r},$$ 
where $r=\sum_{s\in S}r_{s}$. Define the modality-specific true loading matrix and global true loading matrix by $$\mathbf{W}^{(m)}=\big[\mathbf{W}_{123}^{(m)},\cdots,\mathbf{W}_{3}^{(m)}\big]\in\mathbb{R}^{d_m\times r},\quad \mathbf{W}=\left[(\mathbf{W}^{(1)})^\top, (\mathbf{W}^{(2)})^\top, (\mathbf{W}^{(3)})^\top \right]^\top \in\mathbb{R}^{d\times r},$$ 
where $\mathbf{W}_{s}^{(m)}=0$ whenever $m \notin s$.

\begin{theorem}[\textbf{Existence and Identifiability of the Hierarchical Model}]~
\begin{enumerate}[leftmargin=*,align=left]
    \item Suppose the observed multimodal data satisfy $\boldsymbol{x}_i^{(m)} = f^{(m)}\!\left(\mathbf{A}^{(m)}\boldsymbol{u}_i\right)$, where each $f^{(m)}$ is invertible, $\boldsymbol{u}_i \in \mathbb{R}^q$ is a square-integrable random vector with zero mean, and $\mathbf{A}^{(m)} \in \mathbb{R}^{d_m\times q}$ are arbitrary loading matrices that need not satisfy any hierarchical zero pattern. Then this model admits an exact hierarchical decomposition~\eqref{eq:model_hierarchical}, where $\boldsymbol{z}_{s,i}$ has zero mean and identity covariance, and $\mathrm{Cov}(\boldsymbol{z}_{s,i},\boldsymbol{z}_{t,i})=0$ for $s\neq t$.
    \item For the hierarchical decomposition model~\eqref{eq:model_hierarchical}, suppose that each $f^{(m)}$ is invertible and $\boldsymbol{z}_{s,i}$ has zero mean and identity covariance, and $\mathrm{Cov}(\boldsymbol{z}_{s,i},\boldsymbol{z}_{t,i})=0$ for $s\neq t$. For each modality $m$, assume that $\mathbf{W}_{\rm act}^{(m)}=[\mathbf{W}_{s}^{(m)}]_{m\in s,s\in S_*}$ has full column rank. Then the loading matrices and latent variables are identifiable up to structure-wise orthogonal transformations.
\end{enumerate}
\label{thm:hierarchical_framework}
\end{theorem}

Theorem~\ref{thm:hierarchical_framework} shows that any data following $\boldsymbol{x}_i^{(m)} = f^{(m)}\!\left(\mathbf{A}^{(m)}\boldsymbol{u}_i\right)$ admits an exact hierarchical decomposition of the form~\eqref{eq:model_hierarchical}, which is identifiable up to orthogonal transformations under mild conditions. This result also indicates that the proposed hierarchical model is not restrictive. In particular, the nonlinear multimodal factor model with invertible link functions can be represented within this framework after a suitable transformation of the latent variables. Thus, the hierarchical form~\eqref{eq:model_hierarchical} should be viewed as a flexible structural representation rather than a strong modeling assumption.

Relative to existing deterministic matrix-decomposition methods, the main novelty is that identifiability is obtained under weaker structural conditions. For example, \cite{gaynanova2019structural} assumes orthogonality across distinct latent matrices, orthonormality of the latent matrices, and column-wise orthogonality of the loading matrices. The method of \cite{prothero2024data} relaxes the loading condition to full column rank for nonzero loadings, but largely retains the same restrictions on the latent matrices. By contrast, when translated to the deterministic setting, Theorem~\ref{thm:hierarchical_framework} requires only uncorrelated rather than orthogonal latent components , thereby considerably weakening the identifiability conditions.

\subsection{Hierarchical Contrastive Loss}\label{sec:HCLloss}
Given the observed data generated from model~\eqref{eq:model_hierarchical}, our goal is to learn low-dimensional representations that map the raw observations into structure-specific latent spaces. Specifically, for structure $s$ and modality $m$, we define 
$$h_s^{(m)}(\boldsymbol{x}_i^{(m)})=(\mathbf{V}_{s}^{(m)})^\top g^{(m)}(\boldsymbol{x}_i^{(m)}),$$
where $g^{(m)}:\mathcal{X}_{m}\rightarrow\mathbb{R}^{d_m}$ denotes a modality-specific encoder and $\mathbf{V}_{s}^{(m)}\in\mathbb{R}^{d_m\times r_s}$ is used to estimate structure-specific loading matrix. To enforce the hierarchical structure, we set $\mathbf{V}_{s}^{(m)}=0$ whenever $m \notin s$, ensuring that modality $m$ does not contribute to structures with which it is not associated. For notational convenience, define the stacked loading matrix for modality $m$ as 
$\mathbf{V}^{(m)}=\big[\mathbf{V}_{123}^{(m)},\cdots,\mathbf{V}_{3}^{(m)}\big]\in\mathbb{R}^{d_m\times r}$, and the global loading matrix as $\mathbf{V}=[(\mathbf{V}^{(1)})^\top,(\mathbf{V}^{(2)})^\top,(\mathbf{V}^{(3)})^\top]^\top\in\mathbb{R}^{d\times r}$.

To learn these structured representations, we propose a hierarchical contrastive loss that explicitly encourages consistency of the same latent structures across modalities, while suppressing spurious correlations across unrelated structures. Following the basic principle of contrastive learning, the loss is constructed by treating similar representations as positive pairs and dissimilar representations as negative pairs. For each representation $h_{s}^{(m)}(\boldsymbol{x}_i^{(m)})$, we define positive set as the representations of the same structure from the same sample across all modalities, namely $\{h_{s}^{(m')}(\boldsymbol{x}_i^{(m')})\}_{m'=1}^3$, and the negative set as the representations of the same structure from different samples across all modalities, namely $\{h_{s}^{(m')}(\boldsymbol{x}_j^{(m')})\}_{m'=1,j\neq i}^3$. Then the hierarchical contrastive loss is formulated as:
\begin{equation} \label{eq:loss_linear}
    \mathcal{L}_{\rm HCL}(\mathbf{V},\boldsymbol{g})=\frac{1}{2n(n-1)}\sum_{i,j:i\neq j}\phi_{ij}-\frac{1}{2n}\sum_{i=1}^n \phi_{ii}+\frac{\lambda}{4} R(\mathbf{V},\boldsymbol{g}),
\end{equation}
where $\boldsymbol{g}=(g^{(1)}, g^{(2)}, g^{(3)})$, and $\phi_{ij}=\sum_{s\in S}\sum_{m,m'=1}^3\langle  h_{s}^{(m)}(\boldsymbol{x}_i^{(m)}), h_{s}^{(m')}(\boldsymbol{x}_j^{(m')})\rangle$ measures the total similarity between samples $i$ and $j$ across all structures. The regularization term $R(\mathbf{V},\boldsymbol{g})=\|\mathbf{V}^\top\mathbf{V}\|_{\rm F}^2=\sum_{m=1}^3\sum_{s,s'\in S} \|(\mathbf{V}_s^{(m)})^\top\mathbf{V}_{s'}^{(m)}\|_{\rm F}^2$
encourages separation among distinct structures within each modality and regularizes the learned representations to prevent overfitting. This objective therefore encourages representations of the same latent structure from the same sample to be more similar than those from different samples, thereby promoting cross-modal alignment of shared structure while preserving separation across unrelated signals.

When there are more than three modalities, one can define the corresponding hierarchical structure and construct analogous representation functions for each latent component. The main modification relative to loss~\eqref{eq:loss_linear} lies in the definition of $\phi_{ij}$, which must aggregate positive and negative pairs according to the hierarchical structure under consideration.

\begin{remark}[\textbf{Extending to the nonlinear settings}]\label{remark:nonlinear}
    The hierarchical contrastive loss can be applied to general multimodal settings without restrictive assumptions on the data generating process. In particular, the representation $h_{s}^{(m)}$ can be chosen to match the complexity of the data and may be parameterized as a fully nonlinear mapping. More generally, the similarity between $h_{s}^{(m)}(\boldsymbol{x}_i^{(m)})$ and $h_{s}^{(m')}(\boldsymbol{x}_j^{(m')})$ may be measured by $\mathrm{sim}(h_{s}^{(m)}(\boldsymbol{x}_i^{(m)}),h_{s}^{(m')}(\boldsymbol{x}_j^{(m')}))$, where $\mathrm{sim}(\cdot,\cdot)$ denotes a user-specified similarity function. Under this formulation, the regularization term can be generalized accordingly. For example, to encourage separation among different structures within each modality, one may use
    $$R(\boldsymbol{h})=\sum_{i=1}^n\sum_{m=1}^3\sum_{s,s'\in S}\mathrm{sim}(h_{s}^{(m)}(\boldsymbol{x}_i^{(m)}),h_{s'}^{(m)}(\boldsymbol{x}_i^{(m)})).$$
    This term continues to promote disentanglement of distinct latent structures within each modality while helping control overfitting. Furthermore, the proposed loss naturally extends to a softmax-based contrastive loss, analogous to that used in CLIP:
    $$\mathcal{L}_{\rm HCL}(\boldsymbol{h})=-\frac{1}{2n}\sum_{i=1}^n\log \frac{\exp(\phi_{ii}/\tau)}{\frac{1}{n}\sum_{j=1}^n\exp(\phi_{ij}/\tau)}+\lambda R(\boldsymbol{h}),$$
    where $\phi_{ij}=\sum_{s\in S}\sum_{m,m'=1}^3w_s\mathrm{sim}(h_{s}^{(m)}(\boldsymbol{x}_i^{(m)}),h_{s}^{(m')}(\boldsymbol{x}_j^{(m')}))$ with $w_s$ denoting the weight controlling the relative contribution of structure $s$ and $\tau>0$ is a temperature parameter.
\end{remark}

\begin{proposition}\label{pro:loss_svd}
The hierarchical contrastive loss \eqref{eq:loss_linear} can be expressed equivalently as
\begin{equation*}
\mathcal{L}_{\rm HCL}(\mathbf{V},\boldsymbol{g})=\frac{\lambda}{4}\|\mathbf{V} \mathbf{V}^\top-\frac{1}{\lambda}\mathbf{S}_n\|_{\rm F}^2-\frac{1}{4\lambda}\|\mathbf{S}_n\|_{\rm F}^2,
\end{equation*}
where $\mathbf{S}_n=\frac{1}{n-1}\sum_{i=1}^n\boldsymbol{g}(\boldsymbol{x}_i) \boldsymbol{g}(\boldsymbol{x}_i)^\top$, and $\boldsymbol{g}(\boldsymbol{x}_i)=[g^{(1)}(\boldsymbol{x}_i^{(1)})^\top, g^{(2)}(\boldsymbol{x}_i^{(2)})^\top, g^{(3)}(\boldsymbol{x}_i^{(3)})^\top]^\top$.
\end{proposition}

Proposition \ref{pro:loss_svd} shows that, for a fixed encoder $\boldsymbol{g}$, minimizing $\mathcal{L}_{\rm HCL}(\mathbf{V}, \boldsymbol{g})$ reduces to a rank- and structure-constrained approximation of $\mathbf{S}_n$. As shown in Lemma~S3 in Supplementary~S2.2, the tuning parameter $\lambda$ changes only the scale of the minimizing loading matrix through its singular values, but does not affect its left singular subspace. The influence of $\lambda$ on the singular values can always be alleviated by rescaling $\mathbf{V}$ by $\sqrt{\lambda}$. Without loss of generality, we therefore set $\lambda=1$ for simplicity when estimating the loading matrix.

We optimize the hierarchical contrastive loss~\eqref{eq:loss_linear} using gradient descent algorithm, as summarized in Algorithm \ref{alg:GD}, rather than relying on a naive SVD-based solution. This choice is necessary because SVD is applicable only when the encoder function $\boldsymbol{g}$ is known. Even in that setting, a naive SVD approach may recover only the dominant low-rank structure and does not enforce the structural zero constraints required by the hierarchical decomposition. The gradient with respect to $\mathbf{V}$ follows directly from Proposition~\ref{pro:loss_svd}. For the nonlinear encoder $\boldsymbol{g}$, we optimize over a function space $\mathcal{G}$, such as a family of neural network architectures, using gradient-based methods. To improve estimation accuracy, we use the denoised sample covariance $\tilde{\mathbf{S}}_n$ in Algorithm~\ref{alg:GD}. Concretely, denote the eigenvalues of $\mathbf{S}_n$ as $\hat{\sigma}_1\geq\hat{\sigma}_2\geq\cdots\geq\hat{\sigma}_d $. We estimate the noise variance by $\hat{\sigma}_{\epsilon}^2=\frac{1}{d-r}\sum_{j=r+1}^d\hat{\sigma}_j$ and define the denoised sample covariance as $\tilde{\mathbf{S}}_n=\mathbf{S}_n-\hat{\sigma}_{\epsilon}^2\mathbf{I}_d$. To obtain an accurate estimator, we construct a structure-aware initial loading matrix that preserves the hierarchical structure, detailed in Supplementary~S1.

\begin{algorithm}[htb] 
    \caption{Gradient descent algorithm for hierarchical contrastive learning}\label{alg:GD}
    \begin{algorithmic}[1]
        \REQUIRE  Data $\{\boldsymbol{x}_i\}_{i=1}^n$, initial loading matrix $\mathbf{V}_0$ and initial encoder  $\boldsymbol{g}_{\theta_0}(\boldsymbol{x})$  
        \FOR {$t = 1,\cdots,T$}
        \STATE \textbf{Compute} $\mathbf{S}_n^{t-1}=\frac{1}{n-1}\sum_{i=1}^n\boldsymbol{g}_{\theta_{t-1}}(\boldsymbol{x}_i) \boldsymbol{g}_{\theta_{t-1}}(\boldsymbol{x}_i)^\top$ and denoised covariance $\tilde{\mathbf{S}}_n^{t-1}$,
        \STATE \textbf{Update loading matrix:} $\mathbf{V}_t=\mathbf{V}_{t-1}-\eta_t(\mathbf{V}_{t-1} \mathbf{V}_{t-1}^\top-\tilde{\mathbf{S}}_n^{t-1})\mathbf{V}_{t-1}$, where $\eta_t>0$ is the step size,
        \STATE \textbf{Update encoder:} $\theta_t=  \theta_{t-1} - \eta_{g}\frac{\partial}{\partial \theta} \mathcal{L}_{\rm HCL} (\mathbf{V}_{t-1},\boldsymbol{g}_{\theta})$, where $\eta_{g}>0$ is the step size.
        \ENDFOR
        \ENSURE the final loading matrix $\mathbf{V}_{T}$ and encoder $\boldsymbol{g}_{\theta_T}$.
    \end{algorithmic}
\end{algorithm}

\subsection{Downstream Task}
In practice, we are interested in using the learned representations for downstream tasks. Suppose we observe a new data $\{\tilde{\boldsymbol{x}}_i\}_{i=1}^m$ independent of $\{\boldsymbol{x}_i\}_{i=1}^n$, where each $\tilde{\boldsymbol{x}}_i$ follows the hierarchical decomposition model~\eqref{eq:model_hierarchical} with $\tilde{\boldsymbol{x}}_i^{(m)}=f^{(m)}(\mathbf{W}^{(m)} \tilde{\boldsymbol{z}}_i+\tilde{\epsilon}_i^{(m)})~(m=1,2,3),$ 
where $\tilde{\boldsymbol{z}}_{i}, \tilde{\epsilon}_i$ are distributed identically to $\boldsymbol{z}_i,\boldsymbol{\epsilon}_i$, respectively. As an illustrative downstream task, we consider linear regression with response
$$y_i=\boldsymbol{\beta}^{*\top}\tilde{\boldsymbol{z}}_i+\xi_i,$$
where $\boldsymbol{\beta}^*\in\mathbb{R}^r$ is the coefficient and $\xi_i/\sigma_{\xi}$ is sub-Gaussian variable and independent of $\tilde{\boldsymbol{z}}_i$.

Let $\widehat{\mathbf{C}}_{t}=(\mathbf{V}_t \mathbf{V}_t^\top)^{-1}\mathbf{V}_t^\top$ and $\boldsymbol{g}_t(\boldsymbol{x})$ denote the learned recovery operators based on data $\{\boldsymbol{x}_i\}_{i=1}^n$ under model~\eqref{eq:model_hierarchical}. Since the recovered representations are constructed from noisy multimodal observations, applying ordinary least squares directly to $\widehat{\mathbf{C}}_t\boldsymbol{g}_t(\tilde{\boldsymbol{x}}_i)$ leads to an error-in-variables problem and generally yields a biased estimator. To obtain a consistent estimator, we incorporate a debiasing term $\Omega(\boldsymbol{\beta})$ into the least-squares objective \citep{carroll2006measurement}. Moreover, because only a subset of latent structures may be relevant to the downstream task, we also introduce a group Lasso extension to encourage block-wise sparsity and identify predictive latent structures. Specifically, we define
\begin{align*}
   \hat{\boldsymbol{\beta}}=&\arg\min_{\boldsymbol{\beta}}\frac{1}{2m}\sum_{i=1}^m \left[\big(y_i-\boldsymbol{\beta}^\top \widehat{\mathbf{C}}_t\boldsymbol{g}_t(\tilde{\boldsymbol{x}}_i)\big)^2 + \Omega(\boldsymbol{\beta})\right], \\
   \hat{\boldsymbol{\beta}}^{\rm la}=&\arg\min_{\boldsymbol{\beta}}\frac{1}{2m}\sum_{i=1}^m\left[\big(y_i-\sum_{s\in S}\boldsymbol{\beta}_s^\top \widehat{\mathbf{C}}_{s,t}\boldsymbol{g}_t(\tilde{\boldsymbol{x}}_i)\big)^2+ \Omega(\boldsymbol{\beta})\right] + \lambda_m\sum_{s\in S}\|\boldsymbol{\beta}_s\|\ ,
\end{align*}
where $\boldsymbol{\beta}=(\boldsymbol{\beta}_{123}^\top,\cdots,\boldsymbol{\beta}_3^\top)^\top\in\mathbb{R}^r$, $\widehat{\mathbf{C}}_{s,t}\in\mathbb{R}^{r_s\times d}$ denotes the block of rows of  $\widehat{\mathbf{C}}_t$ corresponding to structure $s$, and $\lambda_m$ is the regularization parameter. For the debiased term $\Omega(\boldsymbol{\beta})$, one may take $g^{(m)}=(f^{(m)})^{-1}$ and $\Omega(\boldsymbol{\beta})=- \hat{\sigma}_{\epsilon}^2\boldsymbol{\beta}^\top\widehat{\mathbf{C}}_t\widehat{\mathbf{C}}_t^\top\boldsymbol{\beta}$ if the invertible link functions $f^{(m)}$ are known.

\section{Theoretical Analysis}\label{sec:statistical property}
We now establish the theoretical properties of the proposed HCL framework, including loading recovery from noisy observations and the performance of learned representations in downstream task. Throughout the analysis, we assume that invertible link functions $f^{(m)}$ are known and write the transformed variables $(f^{(m)})^{-1}(\boldsymbol{x^{(m)}})$ still as $\boldsymbol{x}^{(m)}$. This assumption is required for identifiability, since unknown link functions generally prevent unique recovery of the loading matrices from the observed data and therefore create a fundamental barrier to theoretical analysis. It is nevertheless often reasonable in practice, because strong pretrained encoders are available for many modalities and can be regarded as fixed feature maps. In such cases, our method can focus on learning the hierarchical structure without requiring additional training of modality-specific encoders. Even under this assumption, the analysis provides a useful foundation for understanding loading recovery, and downstream performance in hierarchical multimodal representation learning. Extending the theory to unknown or jointly learned link functions is left for future work.

\subsection{Loading Recovery from Noisy Data}
We first study the relationship between the global estimated loading matrix $\mathbf{V}_{t}$ and true loading matrix $\mathbf{W}$. Accurate estimation of $\mathbf{W}$ further enables the recovery of the latent variables $\boldsymbol{z}_i$, which is crucial for downstream applications. Since $\mathbf{W}$ is identifiable only up to orthogonal transformation, we define the optimal alignment of $\mathbf{V}_{t}$ to $\mathbf{W}$ by
$$\mathbf{H}_{t}=\arg\min\limits_{\mathbf{R}\in\mathcal{O}_{r,r}}\|\mathbf{V}_{t} \mathbf{R}-\mathbf{W}\|_{\rm F}.$$
Before presenting the results, we introduce the standard incoherence condition. A rank-$r$ matrix $\mathbf{M}$ with eigen-decomposition $\mathbf{M}=\mathbf{U}\Lambda \mathbf{U}^\top\in\mathbb{R}^{d\times d}$ is said to be $\mu$-incoherent if
$$\|\mathbf{U}\|_{2,\infty}\leq\sqrt{\mu/d}\left\|\mathbf{U}\right\|_{\rm F}=\sqrt{\mu r/d}.$$
Let $\sigma_{\max}^*=\sigma_{1}^{*}\geq\cdots\geq\sigma_{r}^{*}=\sigma_{\min}^*>0$ denote the nonzero eigenvalues of $\mathbf{W}\mathbf{W}^\top$, and define the condition number $\kappa=\sigma_{\max}^*/\sigma_{\min}^*$. We then state the assumptions used for recovery.

\begin{assumption}\label{assum:incoherence_global}
    $\mathbf{W}\mathbf{W}^\top$ is a rank-$r$, $\mu$-incoherent matrix, and $\kappa, \mu$ are bounded.
\end{assumption}
\begin{assumption}\label{assum:distribution_global}
    The coordinates of $\boldsymbol{z}_{s,i}$ and $\boldsymbol{\epsilon}_i^{(m)}/\sigma_{\epsilon}$ are i.i.d. sub-Gaussian with unit variance, and the noise obeys $\sigma^2_{\epsilon}\lesssim\sigma_{\min}^*$.
\end{assumption}

\begin{theorem}\label{thm:GD_global}
Suppose the sample size satisfies $d(d+\log n)\ll n$. Let $\mathbf{U}_r\mathbf{\Lambda}_r\mathbf{U}_r^\top$ denote the top-$r$ SVD of $\tilde{\mathbf{S}}_n$, and initialize the loading matrix as $\mathbf{V}_0 = \mathbf{U}_r (\mathbf{\Lambda}_r)^{\frac{1}{2}}$. Under Assumptions~\ref{assum:incoherence_global} and \ref{assum:distribution_global}, with probability at least $1 - O(\frac{1}{n})$, the loading matrix iterates of Algorithm \ref{alg:GD} with known link functions satisfy
\begin{align*}
\left\| \mathbf{V}_t \mathbf{H}_t - \mathbf{W} \right\| &\lesssim \left( \rho^t+1 \right) \Big[\sqrt{\frac{r+\log n}{n}} + \big(\frac{\sigma_{\epsilon}}{\sqrt{\sigma_{\min}^*}} + \frac{\sigma_{\epsilon}^2}{\sigma_{\min}^*}\big)\sqrt{\frac{d+\log n}{n}}\Big] \left\| \mathbf{W} \right\|, \\
\left\| \mathbf{V}_t \mathbf{H}_t - \mathbf{W} \right\|_{\rm F} &\lesssim \left( \rho^t+1 \right) \Big[\sqrt{\frac{r+\log n}{n}} + \big(\frac{\sigma_{\epsilon}}{\sqrt{\sigma_{\min}^*}} + \frac{\sigma_{\epsilon}^2}{\sigma_{\min}^*}\big)\sqrt{\frac{d+\log n}{n}}\Big] \left\| \mathbf{W} \right\|_{\rm F}, \\
\left\| \mathbf{V}_t \mathbf{H}_t - \mathbf{W} \right\|_{2,\infty} &\lesssim \left( \rho^t+1 \right)\Big[\sqrt{\frac{d(r+\log n)}{nr}} + \big(\frac{\sigma_{\epsilon}}{\sqrt{\sigma_{\min}^*}} + \frac{\sigma_{\epsilon}^2}{\sigma_{\min}^*}\big)\sqrt{\frac{d(d+\log n)}{nr}}\Big] \left\| \mathbf{W} \right\|_{2,\infty}, 
\end{align*}
for all $t\geq 0$, where $1 - \frac{1}{4}\eta\sigma_{\min}^* \leq \rho < 1$, provided that 
$0 < \eta_t \equiv \eta \leq \frac{2}{25 \kappa \sigma_{\max}^*}$.
\end{theorem}

Theorem~\ref{thm:GD_global} establishes recovery of the global loading matrix by providing nonasymptotic error bounds in spectral, Frobenius, and two-infinity norms. In particular, the two-infinity bound ensures that the iterates remain incoherent throughout optimization, thereby ruling out spiky or highly localized errors that cannot be detected by spectral or Frobenius norms alone. Moreover, our results characterize the full matrix error, including singular values, rather than only the associated column space. Compared with existing analyses of multimodal contrastive learning \citep{nakada2023understanding,cai2024contrastive}, which typically provide only subspace-level guarantees, our theory controls a stronger object while retaining a comparable rate up to logarithmic factors. Although HCL imposes hierarchical structural sparsity, this structure does not reduce the order of the unknown nonzero parameters, and hence one should not expect a strictly smaller recovery rate than in unstructured low-rank estimation. Instead, the advantage of HCL lies in its ability to decompose the signal into interpretable hierarchical components while preserving statistically reasonable recovery behavior.

We next consider the recovery of block-wise loading matrices $\mathbf{W}_{s}^{(m)}$. Since the HCL loss explicitly encodes hierarchical structural relationships, we focus on structure-specific rather than modality-specific alignment. For each $s\in S$, define the concatenated estimated and true loading matrices by
$$\mathbf{V}_{s,t}=\left[(\mathbf{V}_{s,t}^{(1)})^\top,(\mathbf{V}_{s,t}^{(2)})^\top,(\mathbf{V}_{s,t}^{(3)})^\top\right]^\top,\quad \mathbf{W}_s=\left[(\mathbf{W}_s^{(1)})^\top,(\mathbf{W}_s^{(2)})^\top,(\mathbf{W}_s^{(3)})^\top\right]^\top.$$
Then the optimal alignment between $\mathbf{V}_{s,t}$ and $\mathbf{W}_{s}$ is defined as 
$$\mathbf{H}_{s,t}=\arg\min\limits_{\mathbf{R}\in\mathcal{O}_{r_s,r_s}}\|\mathbf{V}_{s,t}\mathbf{R}-\mathbf{W}_s\|_{\rm F}.$$
Let $\sigma_{s,1}^{(m)}\geq\cdots\geq\sigma_{s,r_s}^{(m)}>0$ denote the nonzero eigenvalues of the nonzero $\mathbf{W}_s^{(m)}(\mathbf{W}_s^{(m)})^\top$, and define 
$$\sigma_{\max}=\max_{s\in S,m\in[3]}\{\sigma_{s,1}^{(m)}\}, \quad \sigma_{\min}=\min_{s\in S,m\in[3]}\{\sigma_{s,r_s}^{(m)}\},\quad \kappa_{\max}=\sigma_{\max}/\sigma_{\min}.$$
Also write $\max_{s\in S,m\in[3]}\| \mathbf{W}_s^{(m)} \|_{\rm norm}=\|\mathbf{W}_{\max}\|_{\rm norm},$ where the norm denotes the spectral, Frobenius, or two-infinity norm.

\begin{assumption}\label{assum:incoherence_block}
    Each nonzero matrix $\mathbf{W}_s\mathbf{W}_s^\top$ is rank-$r_s$, $\mu_s$-incoherent matrix, and $\kappa_{\max}$ and $\mu_{\max}=\max_{s\in S}\{\mu_s\}$ are bounded.
\end{assumption}

\begin{theorem}\label{thm:GD_block}
Suppose the sample size satisfies $d(d+\log n)\ll n$. Let $\mathbf{V}_0$ be initialized as in Supplementary~S1. Under Assumptions~\ref{assum:distribution_global} and \ref{assum:incoherence_block}, with probability at least $1 - O(\frac{1}{n})$, the loading matrix iterates of Algorithm \ref{alg:GD} with known link functions satisfy
\begin{align*}
    \|\mathbf{V}_{s,t}^{(m)} \mathbf{H}_{s,t} - \mathbf{W}_s^{(m)} \| &\lesssim  (\rho^t+1)\Big[\sqrt{\frac{r+\log n}{n}} + \big(\frac{\sigma_{\epsilon}}{\sqrt{\sigma_{\min}^*}} + \frac{\sigma_{\epsilon}^2}{\sigma_{\min}^*}\big)\sqrt{\frac{d+\log n}{n}}\Big]\| \mathbf{W}_{\max} \|, \\
    \|\mathbf{V}_{s,t}^{(m)} \mathbf{H}_{s,t} - \mathbf{W}_s^{(m)} \|_{\rm F} &\lesssim (\rho^t+1)\Big[\sqrt{\frac{r+\log n}{n}} + \big(\frac{\sigma_{\epsilon}}{\sqrt{\sigma_{\min}^*}} + \frac{\sigma_{\epsilon}^2}{\sigma_{\min}^*}\big)\sqrt{\frac{d+\log n}{n}}\Big] \| \mathbf{W}_{\max}  \|_{\rm F}, \\
    \|\mathbf{V}_{s,t}^{(m)} \mathbf{H}_{s,t} - \mathbf{W}_s^{(m)} \|_{2,\infty} &\lesssim (\rho^t+1)\Big[\sqrt{\frac{d(r+\log n)}{nr}} + \big(\frac{\sigma_{\epsilon}}{\sqrt{\sigma_{\min}^*}} + \frac{\sigma_{\epsilon}^2}{\sigma_{\min}^*}\big)\sqrt{\frac{d(d+\log n)}{nr}}\Big] \| \mathbf{W}_{\max} \|_{2,\infty}, 
\end{align*}
 for all $s\in S, m\in[3]$ and $t>0$, where $1-\frac{1}{4}\sigma_{\min}\leq\rho<1$, provided that $0 < \eta_t \equiv \eta \leq \frac{2}{1369 \kappa_{\max} \sigma_{\max}}$.
\end{theorem}

Theorem~\ref{thm:GD_block} establishes recovery of the full block structure of the hierarchical decomposition, including both the nonzero signal blocks and the zero loading blocks, thereby recovering the prescribed hierarchical sparsity pattern. It is straightforward to verify that $\sigma_{\min}^*\asymp\sigma_{\min}$ and  $\sigma_{\max}^*\asymp\sigma_{\max}$. Thus, the main difference between the two results lies in the incoherence assumption. Theorem~\ref{thm:GD_global} imposes incoherence on the global matrix $\mathbf{W}\mathbf{W}^\top$, whereas Theorem~\ref{thm:GD_block} requires incoherence of the block-wise matrices $\mathbf{W}_{s}\mathbf{W}_s^\top$, which aligns with the respective scopes of the conclusions. In the global analysis, the initialization need not encode structural sparsity, since the full information in the global matrix is retained. By contrast, Theorem~\ref{thm:GD_block} aims to recover the exact structured loading matrices, including their sparsity patterns.

\subsection{Performance in Downstream Task}
The preceding results show that HCL can recover the loading matrix accurately. We now study the performance of the recovered representations in downstream predictions. To obtain accurate estimation of the structured regression parameters, we take the estimated loading matrix to be the concatenation of the structure-specific estimators in Theorem~\ref{thm:GD_block}. Define $\widetilde{\mathbf{H}}_t=\rm{diag}(\mathbf{H}_{123,t},\cdots,\mathbf{H}_{3,t})$, where $\mathbf{H}_{s,t}$ denotes the optimal alignment matrix for structure $s$. Besides the parameter estimation, we also consider the excess risk under the distribution $\mathbb{P}_{\mathbf{W},\boldsymbol{\beta}^*}$ of $(\tilde{\boldsymbol{x}}_i,y_i)$, defined by  
$$\mathcal{E}(\boldsymbol{\beta},\boldsymbol{\beta}^*)=\mathbb{E}_{(\boldsymbol{x},y)\sim\mathbb{P}_{\mathbf{W},\boldsymbol{\beta}^*}}(y-\boldsymbol{\beta}^\top\widehat{\mathbf{C}}_t\boldsymbol{x})^2-\mathbb{E}_{(\boldsymbol{x},y)\sim\mathbb{P}_{\mathbf{W},\boldsymbol{\beta}^*}}(y-\boldsymbol{\beta}^{*\top} \mathbf{C}_*\boldsymbol{x})^2.$$

\begin{theorem}\label{thm:downstream_task}
    Under the conditions of Theorem~\ref{thm:GD_block}, further assume $d+\log m\ll m$. With probability at least $1-O(\frac{1}{n}+\frac{1}{m})$, the ordinary debiased estimator $\hat{\boldsymbol{\beta}}$ satisfies
    \begin{equation*}
    \begin{split}
    \|\widetilde{\mathbf{H}}_{t}^\top\hat{\boldsymbol{\beta}}-
    \boldsymbol{\beta}^*\|\lesssim& (\rho^t+1)\Big[\sqrt{\frac{r+\log n}{n}} + \big(\frac{\sigma_{\epsilon}}{\sqrt{\sigma_{\min}^*}} + \frac{\sigma_{\epsilon}^2}{\sigma_{\min}^*}\big)\sqrt{\frac{d+\log n}{n}}\Big] \\
    &+  (\|\boldsymbol{\beta}^{*}\| +\sigma_\xi)\|\sqrt{\frac{r+\log m}{m}},\\
        \mathcal{E}(\hat{\boldsymbol{\beta}},\boldsymbol{\beta}^*)
    \lesssim& (\|\boldsymbol{\beta}^*\|^2+1)(\rho^t+1)^2\Big[\frac{r+\log n}{n} + \big(\frac{\sigma_{\epsilon}^2}{\sigma_{\min}^*} + \frac{\sigma_{\epsilon}^4}{\sigma_{\min}^{*2}}\big)\frac{d+\log n}{n}\Big] \\
    &+  (\|\boldsymbol{\beta}^{*}\|^2 +\sigma_\xi^2)\|\frac{r+\log m}{m}.
    \end{split}
    \end{equation*}
\end{theorem}

\begin{theorem}\label{thm:downstream_task_GL}
    Under the conditions of Theorem~\ref{thm:downstream_task}, set $\lambda_m=C(\rho^t+1)\Big[\sqrt{\frac{r+\log n}{n}} + \big(\frac{\sigma_{\epsilon}}{\sqrt{\sigma_{\min}^*}} + \frac{\sigma_{\epsilon}^2}{\sigma_{\min}^*}\big)\sqrt{\frac{d+\log n}{n}}\Big] + C(\|\boldsymbol{\beta}^{*}\| +\sigma_\xi)\|\sqrt{\frac{r+\log m}{m}},$ where C is a universal constant. Then with probability at least $1-O(\frac{1}{n}+\frac{1}{m})$, the group lasso estimator $\hat{\boldsymbol{\beta}}^{\rm la}$ satisfies 
    \begin{equation*}
    \begin{split}
     \max_{s\in S}\|\mathbf{H}_{s,t}^\top\hat{\boldsymbol{\beta}}_s^{\rm la}-
    \boldsymbol{\beta}_s^*\|\lesssim& (\rho^t+1)\Big[\sqrt{\frac{r+\log n}{n}} + \big(\frac{\sigma_{\epsilon}}{\sqrt{\sigma_{\min}^*}} + \frac{\sigma_{\epsilon}^2}{\sigma_{\min}^*}\big)\sqrt{\frac{d+\log n}{n}}\Big] \\
    &+  (\|\boldsymbol{\beta}^{*}\| +\sigma_\xi)\|\sqrt{\frac{r+\log m}{m}},\\
        \mathcal{E}(\hat{\boldsymbol{\beta}}^{\rm la},\boldsymbol{\beta}^*)
    \lesssim& (\|\boldsymbol{\beta}^*\|^2+1)(\rho^t+1)^2\Big[\frac{r+\log n}{n} + \big(\frac{\sigma_{\epsilon}^2}{\sigma_{\min}^*} + \frac{\sigma_{\epsilon}^4}{\sigma_{\min}^{*2}}\big)\frac{d+\log n}{n}\Big] \\
    &+  (\|\boldsymbol{\beta}^{*}\|^2 +\sigma_\xi^2)\|\frac{r+\log m}{m}.
    \end{split}
    \end{equation*}
\end{theorem}

These two theorems provide non-asymptotic error bounds for downstream estimation and prediction based on the learned representations. Theorem~\ref{thm:downstream_task} shows that the debiased estimator attains controlled parameter estimated error and excess risk after orthogonal alignment, with the total error determined by two sources: the representation recovery error inherited from Theorem~\ref{thm:GD_block} and the additional sampling noise from the downstream regression problem. Theorem~\ref{thm:downstream_task_GL} strengthens this conclusion by giving block-wise control for the group Lasso estimator, thereby enabling identification of task-relevant latent structures rather than only overall predictive accuracy. Relative to the unpenalized debiased estimator, the main novelty of the group-penalized result is that it promotes structured shrinkage, so blocks corresponding to irrelevant modalities or latent structures are driven toward zero. In applications such as electronic health records, where modalities may be missing, redundant, or irrelevant, this property offers useful guidance for modality screening, efficient data collection, and downstream interpretation.

\section{Simulation Studies}\label{sec:simulation}

We conduct simulation studies to validate the theoretical results. Under identity link functions, we examine both representation learning and downstream task performance.

\subsection{Representation Learning Results}\label{sec:simu_HCL}
In this section, we evaluate the performance of three methods: (i) the top-$r$ SVD of the sample covariance matrix (Naive-SVD), (ii) the initial structure-aware SVD described in Supplementary~S1 (HCL-SVD), and (iii) the gradient-based optimization Algorithm~\ref{alg:GD} (HCL-GD), initialized as in Supplementary~S1. For the gradient-based method, we fix the regularization parameter at $\lambda=1$ and initialize the learning rate at $10^{-4}$, reducing it by a factor of $10$ every $10$ epochs to promote stable convergence. The algorithm is terminated when the change in the loss falls below $10^{-6}$. 

The latent variables $\boldsymbol{z}_{s,i}$ are generated independently from the multivariate normal distribution $N(\mathbf{0},\mathbf{I}_{r/7})$ with $r_s=r/7$. For each modality $m$ and structure $s$, the left and right singular vector matrices $\mathbf{U}_s^{(m)}$ and $\mathbf{V}_s^{(m)}$ are sampled uniformly from the set of orthogonal matrices $\mathcal{O}_{d_m\times r_s}$. The diagonal entries of $\mathbf{\Sigma}_{s}^{(m)}$ are drawn independently from the uniform distribution ${\rm U}(0.5,1.5)$ and sorted in descending order. The loading matrices are then constructed as $\mathbf{W}_{s}^{(m)}=\mathbf{U}_s^{(m)}\mathbf{\Sigma}_{s}^{(m)}(\mathbf{V}_s^{(m)})^\top$. The noise terms $\boldsymbol{\epsilon}_i^{(m)}$ in \eqref{eq:model_hierarchical} follow $N(0,c\mathbf{I}_{d_m})$, where $c$ controls the signal-to-noise ratio (SNR) of the data generating process. For the global loading matrix, as well as each nonzero block-wise loading matrix, recovery accuracy is evaluated using $\mathrm{Err}(\mathbf{V},\mathbf{W})=\|\mathbf{V}\mathbf{V}^\top-\mathbf{W}\mathbf{W}^\top\|_{\rm F}$ (and its block-wise counterpart). Because Naive-SVD does not preserve the hierarchical structure, the estimated and true loading matrices are first aligned before evaluation. Specifically, we use the structured projection matrix $\mathbf{V}\mathbf{H}$, where $\mathbf{H}=\arg\min\limits_{\mathbf{R}\in\mathcal{O}_{r,r}}\|\mathbf{V} \mathbf{R}-\mathbf{W}\|_{\rm F}$.

\begin{figure}[htb]
    \centering
    \includegraphics[width=0.85\linewidth]{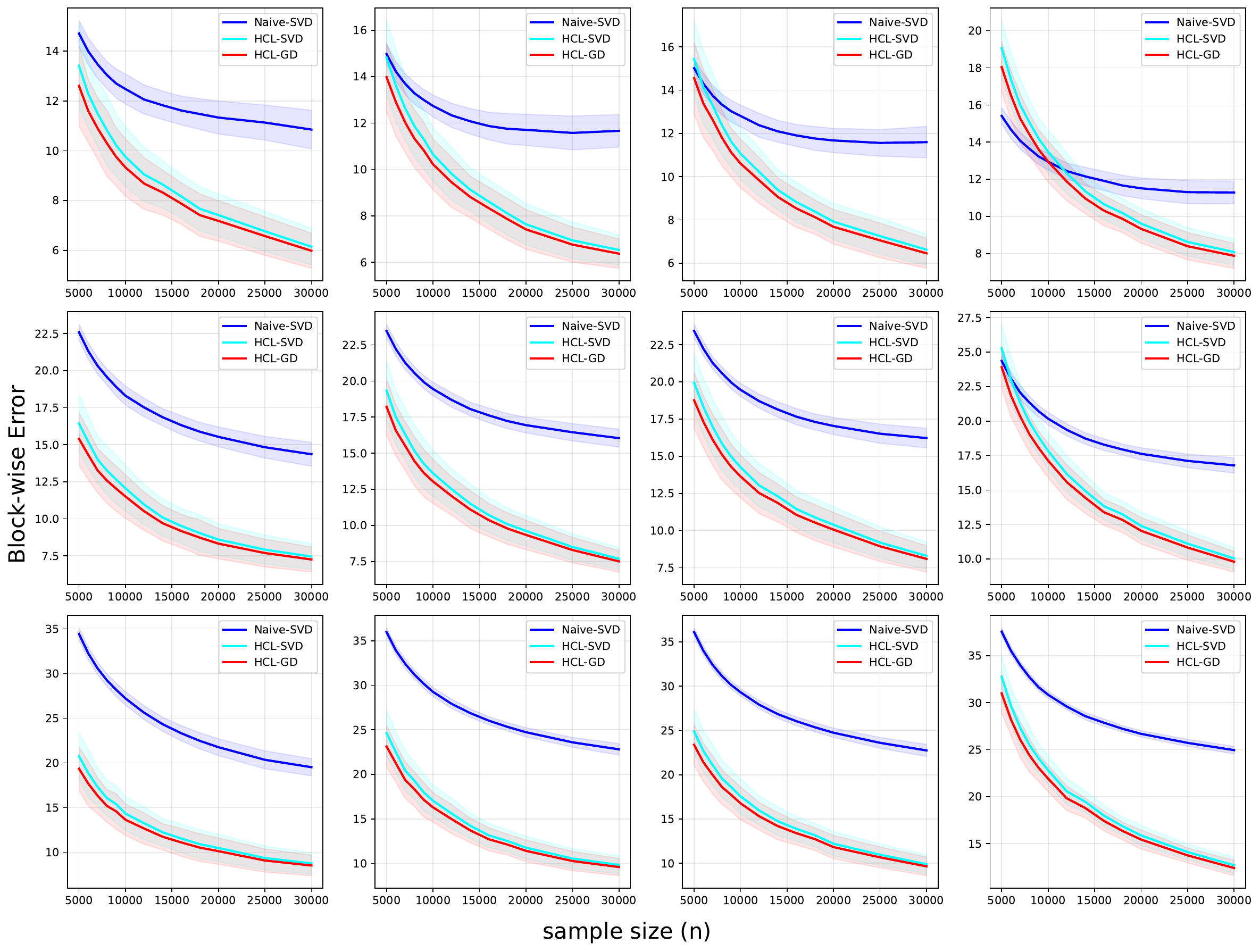}\\
    \caption{Block matrices error of the hierarchical contrastive learning (HCL) framework as a function of the unlabeled sample size for modality dimensions $(300,500,800)$. The twelve subplots are organized by modality (rows) and latent structure (columns).}
    \label{fig:feature_recovery}
\end{figure}

\begin{figure}[htb]
    \centering
    \includegraphics[width=0.3\linewidth]{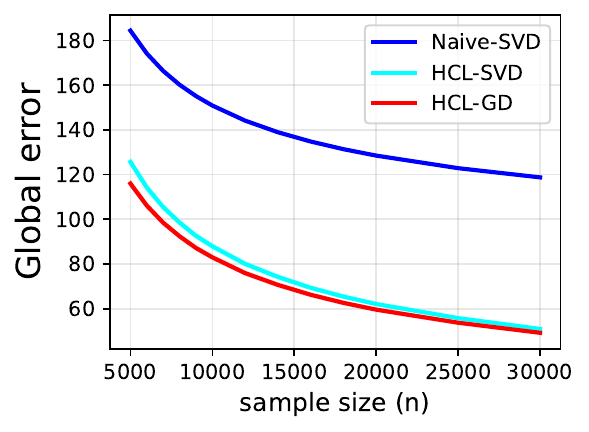}
    \includegraphics[width=0.3\linewidth]{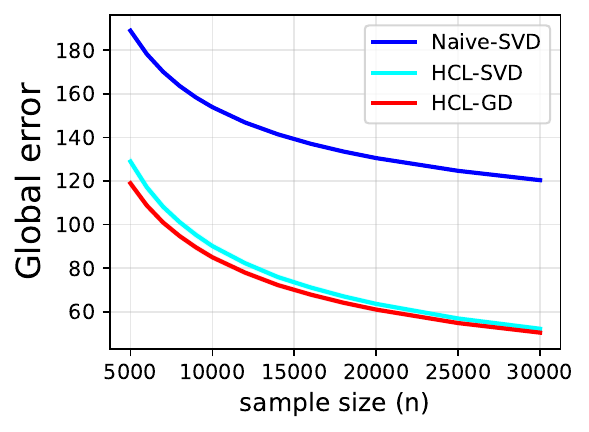}
    \includegraphics[width=0.3\linewidth]{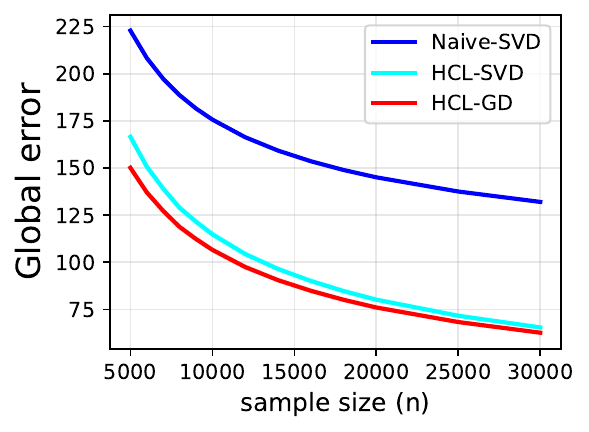}
    \caption{Global matrix error of the hierarchical contrastive learning (HCL) framework versus unlabeled sample size. The three panels correspond, from left to right, to modality dimensions (500, 500, 500), (300, 500, 800), and (800, 800, 800)}
    \label{fig:feature_recovery_global}
\end{figure}

We fix the latent dimension at $r=70$ and the noise level at $c=10$, and vary the sample size over $\{5, 10, 15, 20, 25, 30\}\times 10^3$ to study loading recovery under different modality dimensions. For each setting, performance is evaluated over $100$ replications, and we report the mean and standard error of the recovery metric. Results for different modality dimensions as a function of the sample size $n$ are shown in Figure~\ref{fig:feature_recovery} and \ref{fig:feature_recovery_global}. 
 
It can be seen that both the block-wise and global error metrics decrease as the sample size increases, which is consistent with the theoretical results. In addition, the gradient-based algorithm consistently outperforms both the Naive-SVD baseline and the HCL-SVD method. The first comparison indicates that the naive SVD solution implied by the Eckart–Young–Mirsky theorem is not optimal for estimating loading matrices under the proposed hierarchical structure, while the second agrees with the theoretical prediction that the estimation error decreases as the number of iterations increases. Additional simulation results for other modality dimensions are reported in Supplementary~S5.1, where similar conclusions are observed.

\subsection{Downstream Task Results}
\begin{figure}[t]
    \centering
    \includegraphics[width=\textwidth]{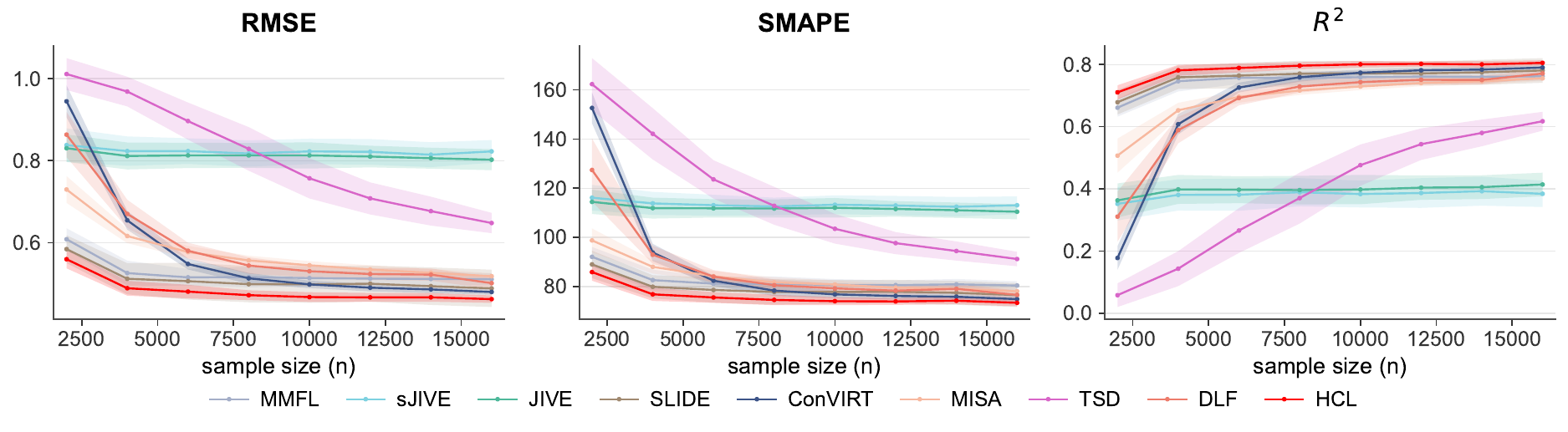}
    \caption{Downstream prediction performance of different methods under varying sample size $n$ with noise level $c = 0.1$, $\sigma_{\xi} = 0.1$.}
    \label{fig:downstream_simu}
\end{figure}

For downstream evaluation, we generate the data for HCL under the same setting as in Section~\ref{sec:simu_HCL}, except that the noise level is set to $c=0.1$. We consider the in-domain setting, in which the latent variables $\tilde{\boldsymbol{z}}_i$ follow the same distribution as $\boldsymbol{z}_i$. The regression coefficient $\boldsymbol{\beta}^*$ is sampled from the unit sphere, and response noise level is set to $\sigma_{\xi}=0.1$. Because the gradient-based HCL method performs better than the SVD-based version and extends naturally to nonlinear settings, we use the gradient-based procedure for downstream prediction. We compare the proposed method with a range of related approaches, including deterministic matrix factorization methods including JIVE \citep{lock2013joint}, SLIDE \citep{gaynanova2019structural}, sJIVE \citep{palzer2022sjive}, MMFL \citep{mao2026supervised}, as well as deep multimodal learning methods including MISA \citep{hazarika2020misa}, ConVIRT \citep{zhang2022contrastive},  DLF \citep{wang2025dlf}, TSD \citep{meng2026tri}. To evaluate downstream performance, we report RMSE (root mean squared error), SMAPE (symmetric mean absolute percentage error), and $R^2$, with results averaged over 100 replications. Figure~\ref{fig:downstream_simu} displays prediction performance for all methods as a function of the sample size $n$. 

In Figure~\ref{fig:downstream_simu}, as the sample size increases, the proposed method achieves lower prediction error, as measured by RMSE and SMAPE, together with higher $R^2$. Compared with methods based on a simple shared-versus-private decomposition, such as JIVE and sJIVE, approaches that explicitly accommodate partially shared structure, including SLIDE, MMLF, and HCL, achieve better downstream performance. This pattern highlights the importance of hierarchical decomposition in multimodal data. In addition, compared with deep multimodal learning methods, whose performance is less stable at small sample sizes, HCL is less sensitive to sample size and performs more favorably across the full range of sample size. Overall, HCL consistently outperforms the competing methods on all three metrics, demonstrating the advantage of the proposed framework. We also report results under an alternative setting with $c=0.5$ and $\sigma_{\xi}=0.2$ in Supplementary~S5.2, which shows a similar  pattern.

We further conduct additional simulation experiments in Supplementary~S5.2 to examine the estimation error of the structural parameters for both the debiased estimator and the group Lasso estimator, while varying the representation-learning and downstream sample sizes separately. The results show that the estimation error decreases as either sample size increases, and that the group Lasso estimator successfully identifies the structures relevant to the downstream task.

\section{Application to Electronic Health Record Studies}\label{sec:real_data}
To evaluate the practical utility of the proposed HCL framework, we consider a multimodal electronic health record (EHR) application based on the MIMIC-IV \citep{johnson2020mimic} database, a large de-identified clinical dataset. We construct three modalities from the longitudinal records: discrete clinical codes, continuous physiological measurements derived from laboratory events, and unstructured clinical notes. To define a consistent multimodal cohort and retain longitudinal disease progression, we restrict the analysis to patients with at least two recorded visits and available information in all three modalities, yielding a final analytic cohort of 108{,}927 patients. Because the raw sequences are irregular and of varying length, each modality is mapped to a fixed-dimensional patient-level representation using modality-specific sequence modeling and natural language processing pipelines. The details on cohort construction, code mapping, modality-specific embeddings, and preprocessing are given in Supplementary~S6.1.

We consider three downstream tasks representing distinct aspects of patient risk: $30$-day readmission prediction, defined as whether a patient is readmitted within $30$ days after discharge from the penultimate hospitalization; prediction of the length of stay of the next hospitalization, as a proxy for subsequent healthcare utilization; and one-year mortality prediction, defined as whether the patient dies within one year after the last recorded visit. For the two classification tasks, we report AUC (area under the ROC curve), AUPRC (area under the precision--recall curve), ACC (accuracy), and F1 score; for length-of-stay prediction, we report MAE (mean absolute error), RMSE, SMAPE, and $R^2$.

Here we apply the fully nonlinear version of HCL to accommodate complex data, using neural-network-based encoders $h_{s,\theta}(\boldsymbol{x})$ and incorporating the regularization term with inner product from Remark~\ref{remark:nonlinear} into the loss function \eqref{eq:loss_linear}. We compare HCL with the same baselines as in the simulation study, using identical patient-level modality inputs, multilayer perceptron prediction heads, and data splits (70\% training, 10\% validation, and 20\% test) for all methods. To ensure a fair comparison and isolate the benefits of hierarchical modeling, we align the latent dimensions of all baselines with the corresponding structural dimensions learned by HCL (see Supplementary~S6.3).

\begin{figure}[tbp]
    \centering
    \includegraphics[width=1\linewidth]{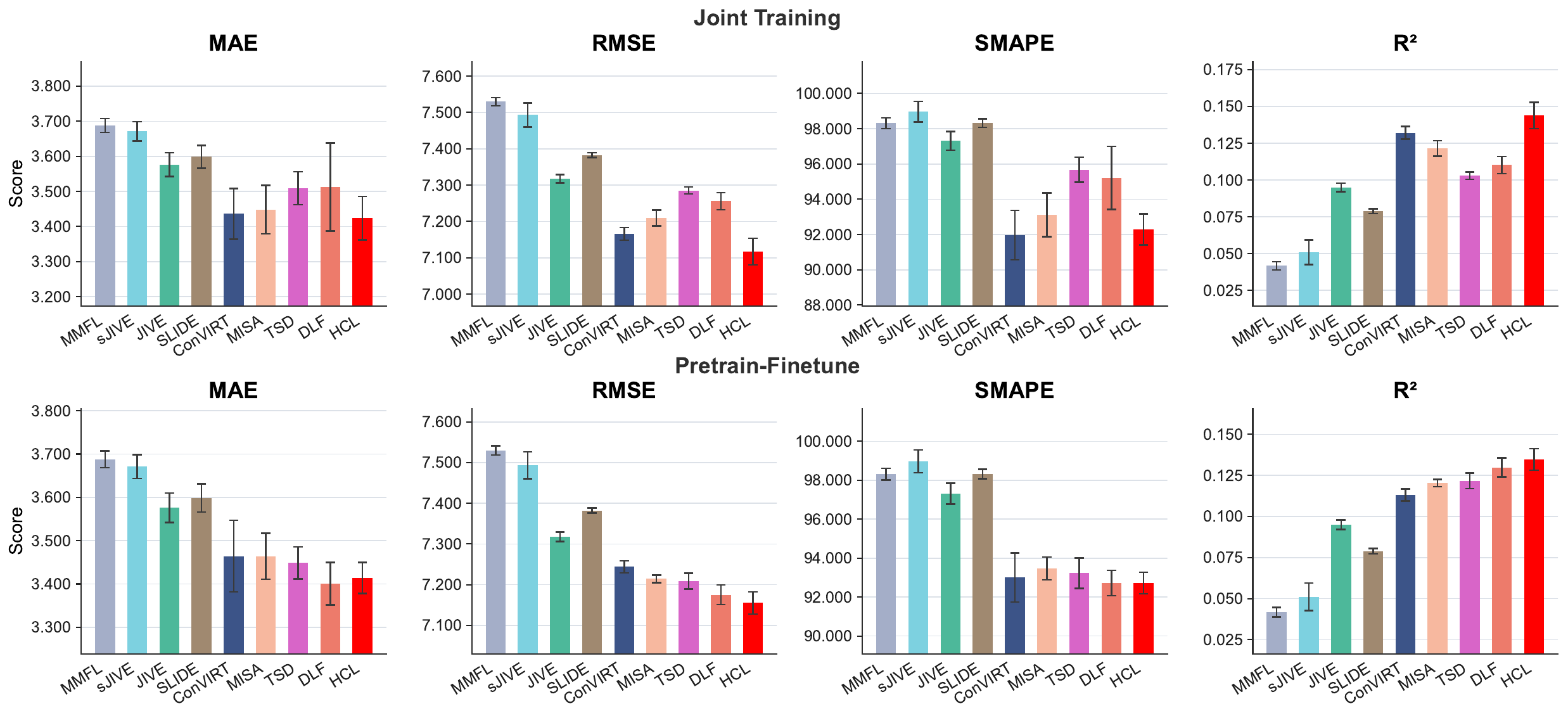}
    \caption{Comparison of methods for next-visit length-of-stay prediction under joint training and pretrain--finetune. Results are averaged over 50 random splits.}
    \label{fig:los}
\end{figure}

\begin{figure}[tbp]
    \centering
    \includegraphics[width=1\linewidth]{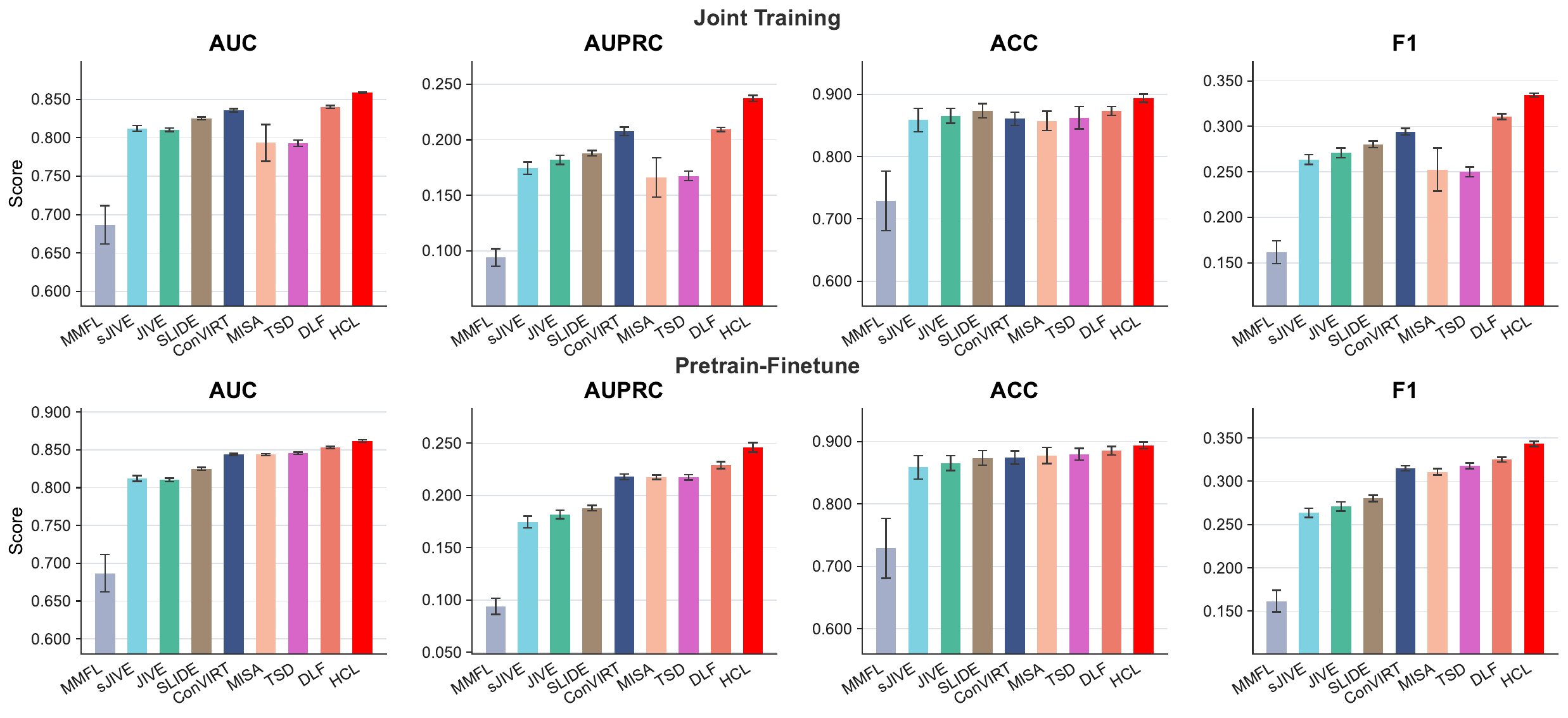}
    \caption{Comparison of methods for one-year mortality prediction under joint training and pretrain--finetune. Results are averaged over 50 random splits.}
    \label{fig:1ydeath}
\end{figure}

Both joint training and pretrain-finetune strategies are considered. In joint training, the representation module and prediction head are optimized end-to-end for each task. In pretrain-finetune, the representation model is first pretrained in a task-agnostic manner and then fine-tuned jointly with the prediction head. All models are optimized using the Adam optimizer. To optimize the representational capacity for each hierarchical level, we employ the Optuna framework to adaptively search for the optimal latent dimension list ($r$-list) across the 7 hierarchical structures, based on validation set performance. The search space, training hyperparameters (e.g., learning rates, early stopping), and model selection criteria are detailed in Supplementary~S6.3. Results in Figures~\ref{fig:mimic_readmission}-\ref{fig:1ydeath} are averaged over 50 random splits.

\begin{figure}[tbp]
    \centering
    \includegraphics[width=1\linewidth]{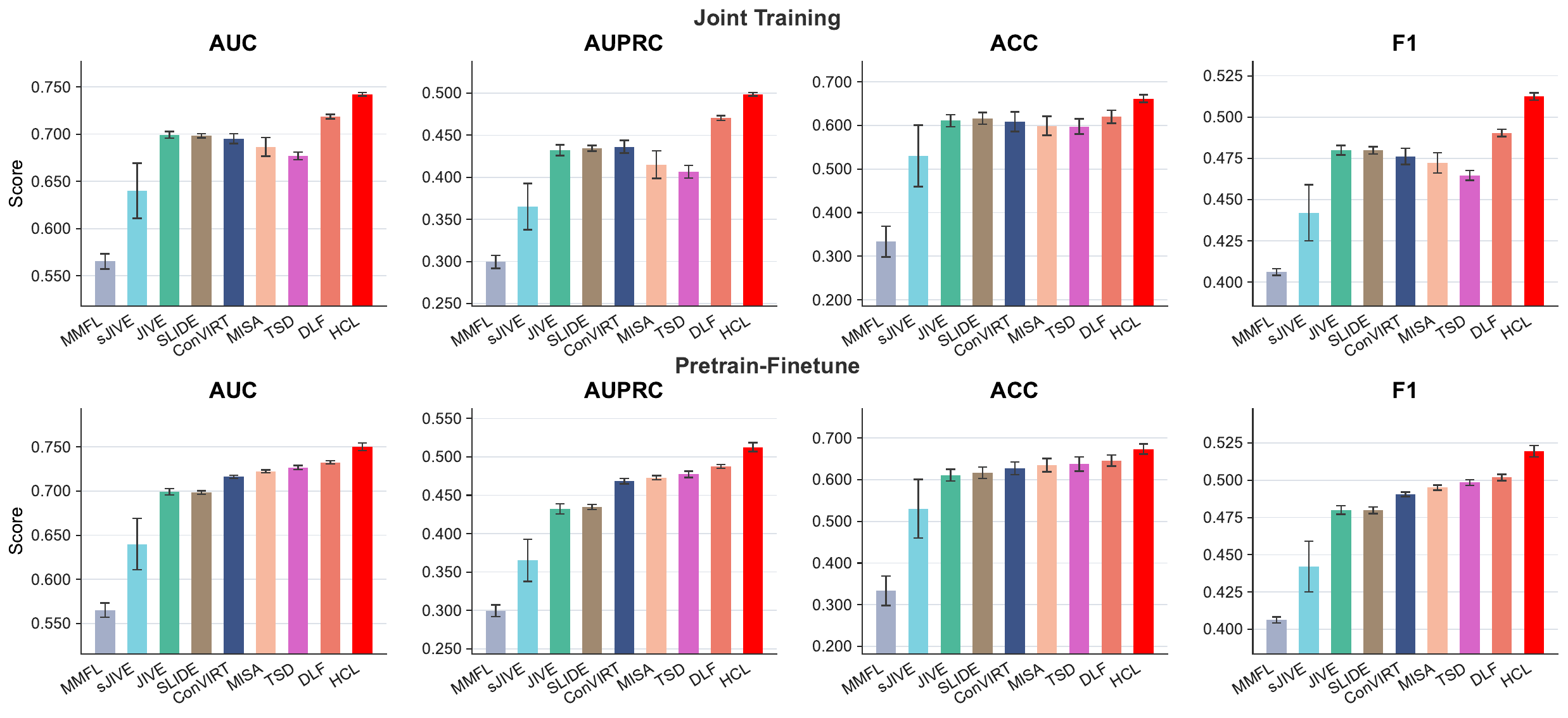}
    \caption{Comparison of methods for 30-day readmission prediction under joint training and pretrain--finetune. Results are averaged over 50 random splits.}
    \label{fig:mimic_readmission}
\end{figure}

The results show that HCL consistently improves performance across the three clinical tasks, although the magnitude and form of improvement vary by outcome. For $30$-day readmission prediction (Figure~\ref{fig:mimic_readmission}), HCL achieves the best performance across all metrics under both joint training and pretrain-finetune, suggesting that short-term readmission risk depends on multimodal signals with dependence structure too complex to be captured by a single shared component or a simple shared-versus-private split. For next-visit length-of-stay prediction (Figure~\ref{fig:los}), HCL again performs best under joint training across all regression metrics and remains best in $R^2$ and competitive in RMSE under pretrain-finetune, although its advantage is smaller for MAE and SMAPE. For one-year mortality prediction (Figure~\ref{fig:1ydeath}), HCL achieves the best overall performance under both training strategies, with particularly strong gains in AUPRC and F1 score. Taken together, these findings show that HCL yields stable improvements across tasks and is especially effective when downstream outcomes depend on complex multimodal structure. The small difference in downstream performance between pretrain--finetune and joint training suggests that much of the hierarchical structure exploited by HCL is already present in the multimodal data and can be learned with limited supervision from response labels. This finding underscores the potential advantage of HCL in unsupervised or weakly supervised settings.

\begin{figure}[tbp]
    \centering
    \includegraphics[width=1\linewidth]{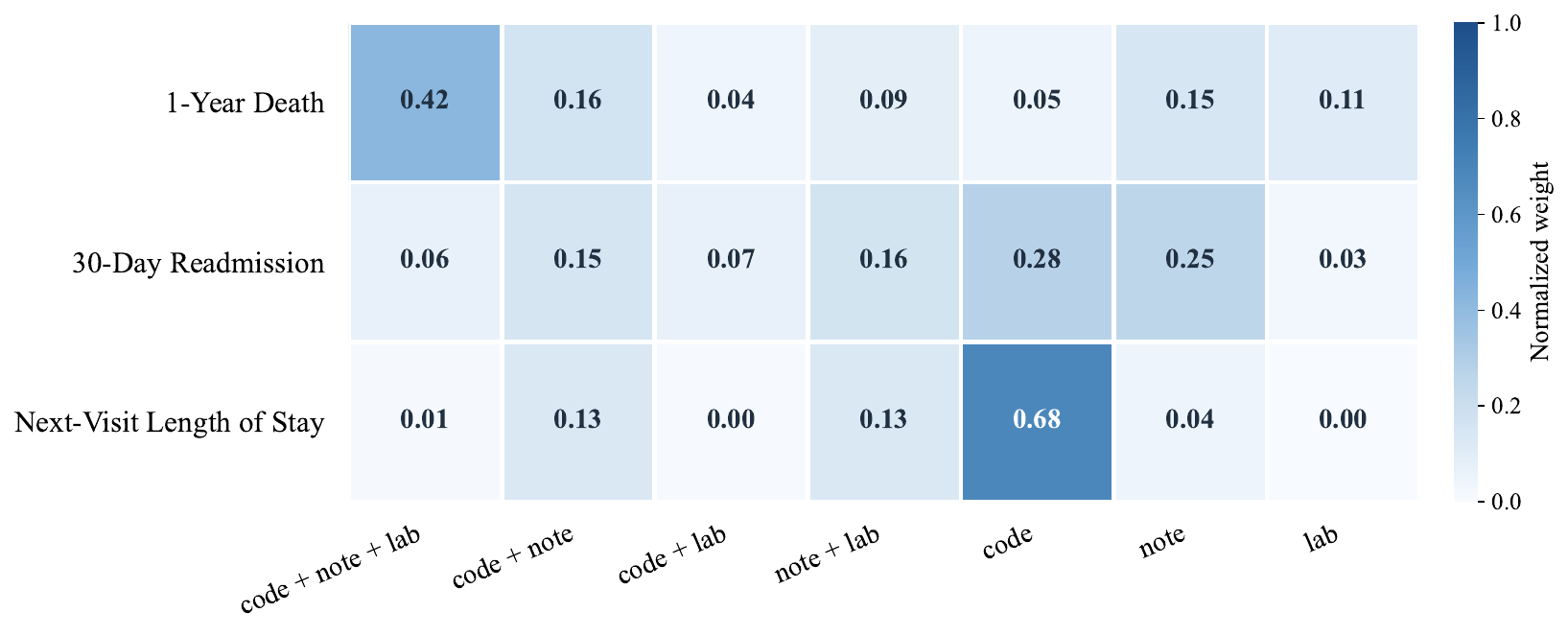}
    \caption{Normalized importance weights of latent structures for the three downstream tasks under HCL. Results are averaged over 5 independent runs.}
    \label{fig:structure_importance_weight_paper}
\end{figure}

To further quantify the contribution of different latent structures to downstream prediction, Figure~\ref{fig:structure_importance_weight_paper} reports the normalized importance weights across the three tasks under the joint training setting, where the latent dimensions are fixed to be the same across structures ($r_s = 100$ for all $s$) to ensure comparability across structures and avoid confounding effects due to differing representation capacities. We assess the importance of each structure through an ablation procedure, in which the representation block corresponding to a given structure is masked and the resulting deterioration in predictive performance is measured. The raw importance score is defined as the decrease in AUC for classification tasks and the increase in MSE for the regression task, relative to the full model, and is then normalized across all candidate structures. It can be seen that the importance of globally shared, partially shared, and modality-specific components varies substantially by outcome. The globally shared component is most important for one-year mortality prediction ($0.42$), whereas the code-only and note-only components receive the largest weights for 30-day readmission prediction ($0.28$ and $0.25$, respectively). For next-visit length-of-stay prediction, the code-only component is dominant ($0.68$), whereas all remaining components receive considerably smaller weights. Together with the predictive results, these findings underscore the value of modeling multiple levels of shared structure in multimodal EHR data. Additional visualization of the learned latent representations, illustrating the separation of different latent structures across downstream tasks, is provided in Supplementary~S6.2.

\section{Discussion}\label{sec:discussion}
We proposed HCL, a multimodal representation learning framework that explicitly captures globally shared, partially shared, and modality-specific information. By combining a hierarchical latent-variable model with a structured contrastive objective, HCL yields identifiable hierarchical decompositions, consistent recovery of loading matrices, and nonasymptotic guarantees for downstream linear regression.

The framework is motivated by the observation that multimodal data rarely follow a purely shared-versus-specific pattern. In many applications, important factors are shared by only a subset of modalities, and explicitly modeling this intermediate level of sharing can improve both interpretability and downstream performance. In applications such as EHR data, this distinction helps clarify whether predictive information is broadly supported across modalities or driven by a single source or subset of sources.

Future work can extend HCL in several important directions. Theoretically, it would be valuable to move beyond known invertible link functions and accommodate unknown or noninvertible links as well as model misspecification. Methodologically, scalable procedures are needed for selecting hierarchical structures, latent dimensions, and regularization parameters, particularly in high-modality settings where the number of candidate partially shared components can grow quickly. Practically, extending the framework to missing modalities, asynchronous measurements, and more general downstream outcomes, such as survival and longitudinal responses, would greatly broaden its applicability, especially in biomedical problems. More broadly, these directions suggest a fruitful integration of multimodal contrastive learning with structured statistical modeling, combining the flexibility of modern representation learning with stronger interpretability, identifiability, uncertainty quantification, and principled support for downstream scientific decision-making.

\noindent \textbf{Acknowledgments\,}This work was partially supported by the National Natural Science Foundation of China No.12571298, Fundamental Research Funds for the Central Universities and Fundamental and Interdisciplinary Disciplines Breakthrough Plan of the Ministry of Education of China(JYB2025XDXM612).

\bibliography{reference.bib}

\end{sloppypar}
\end{document}